\documentclass{article}

\PassOptionsToPackage{numbers, compress}{natbib}

\usepackage[final]{neurips_2022}

\usepackage[utf8]{inputenc} 
\usepackage[T1]{fontenc}    
\usepackage{hyperref}
\hypersetup{colorlinks, urlcolor=blue}
\usepackage{url}            
\usepackage{booktabs}       
\usepackage{amsfonts}       
\usepackage{nicefrac}       
\usepackage{microtype}      
\usepackage{xcolor}         
\usepackage{graphicx}
\usepackage{float}
\usepackage{subfigure} 
\usepackage{caption}
\usepackage{natbib}
\usepackage{amsfonts,amssymb}
\usepackage[namelimits]{amsmath} %
\usepackage{amssymb}             %
\usepackage{amsfonts}            %
\usepackage{mathrsfs} 
\usepackage{tabularx}
\usepackage{wrapfig}
\usepackage{verbatim}
\usepackage{dsfont}
\usepackage{tablefootnote}
\usepackage[stable]{footmisc}
\usepackage{multicol}
\usepackage{multirow}
\usepackage{cleveref}

\usepackage{color}

\newcommand{\red}[1]{{\color{red}#1}}
\newcommand{\blue}[1]{{\color{blue}#1}}

\definecolor{jiapeng}{rgb}{1.0,0.0,0.0}

\definecolor{bi}{rgb}{1.0,0.75,0.0}

\definecolor{lev}{rgb}{0.0,0.5,0.0}

\definecolor{justus}{rgb}{0.6,0.3,0.9}

\definecolor{matthias}{rgb}{0.9,0.,0.5}

\iftrue
	\definecolor{jtcolor}{RGB}{0,0,255}
	\newcommand\JT[1] {\emph{\textcolor{jtcolor}{JT: #1}}}

	\definecolor{todocolor}{RGB}{255,0,00}
	\newcommand\TODO[1] {\PackageWarning{}{Unprocessed todo}\emph{\textcolor{todocolor}{TODO: #1}}}
	\newcommand\rev[1] { {}{#1} }
\else 
	\newcommand\JT[1] {}
	\newcommand\TODO[1] {}
\fi 

\DeclareMathOperator{\VCA}{VCA}
\DeclareMathOperator{\VSA}{VSA}
\DeclareMathOperator{\BN}{BN}
\DeclareMathOperator{\PTB}{PTB}
\DeclareMathOperator{\enc}{enc}
\DeclareMathOperator{\dec}{dec}

\DeclareMathOperator{\PAB}{PAB}
\DeclareMathOperator{\glo}{glo}
\DeclareMathOperator{\FPS}{FPS}
\DeclareMathOperator{\CD}{CD}
\DeclareMathOperator{\FNC}{FNC}

\title{Neural Shape Deformation Priors}

\author{
  Jiapeng Tang$^1$ \quad Lev Markhasin$^2$ \quad Bi Wang$^2$ \quad Justus Thies$^3$ \quad Matthias Nie{\ss}ner$^1$ \\
    \\
     $^{1}$ Technical University of Munich \quad
     $^{2}$ Sony Europe RDC Stuttgart \\ 
     $^{3}$ Max Planck Institute for Intelligent Systems, Tübingen, Germany \\
     \url{https://tangjiapeng.github.io/projects/NSDP/} \\
}

\begin{document}

\maketitle

\begin{figure}[h!]
	\centering
	\vspace{-0.5cm}
	\includegraphics[width=\linewidth]{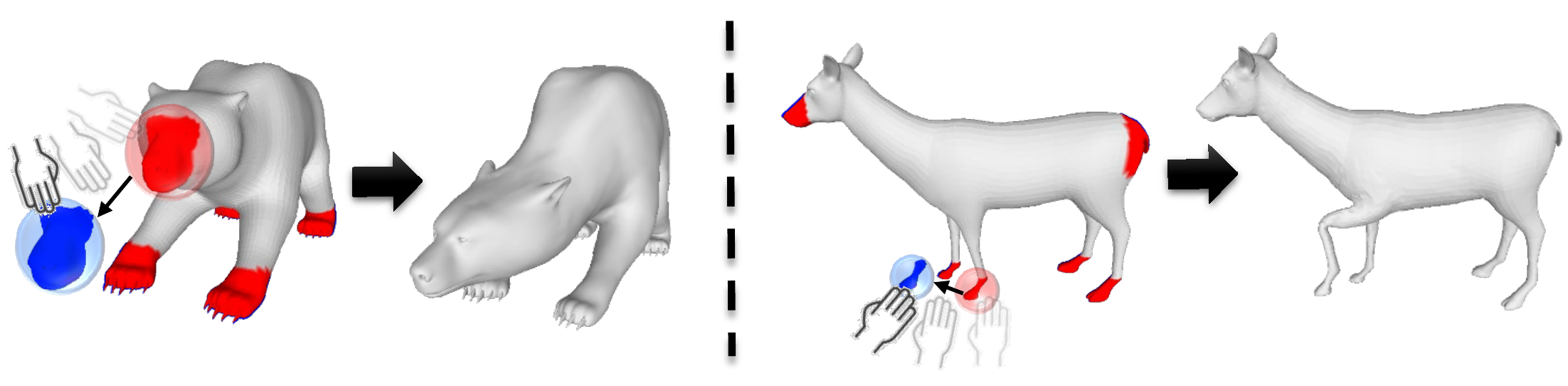}
	\caption{
	%
	%
	Neural shape deformation priors allow for intuitive shape manipulation of existing source meshes. A user can create novel shapes by dragging handles (\red{red} circles) defined on the region of interest (\red{red} regions) to desired locations (\blue{blue} circles).
	}
    \label{fig:teaser}
\end{figure}

\begin{abstract}
We present \textit{Neural Shape Deformation Priors}, a novel method for shape manipulation that predicts mesh deformations of non-rigid objects from user-provided handle movements.
State-of-the-art methods cast this problem as an optimization task, where the input source mesh is iteratively deformed to minimize an objective function according to hand-crafted regularizers such as ARAP~\cite{sorkine2007rigid}.
In this work, we learn the deformation behavior based on the underlying geometric properties of a shape, while leveraging a large-scale dataset containing a diverse set of non-rigid deformations.
Specifically, given a source mesh and desired target locations of handles that describe the partial surface deformation, we predict a continuous deformation field that is defined in 3D space to describe the space deformation.
%
To this end, we introduce transformer-based deformation networks that represent a shape deformation as a composition of local surface deformations.
It learns a set of local latent codes anchored in 3D space, from which we can learn a set of continuous deformation functions for local surfaces.
Our method can be applied to challenging deformations and generalizes well to unseen deformations.
%
%
We validate our approach in experiments using the DeformingThing4D dataset, and compare to both classic optimization-based and recent neural network-based methods.

\end{abstract}
\section{Introduction}
\label{SecIntro}
Editing and deforming 3D shapes is a key component in animation creation and computer aided design pipelines.
Given as little user input as possible, the goal is to create new deformed instances of the original 3D shape which look natural and behave like real objects or animals.
The user input is assumed to be very sparse, such as  vertex handles that can be dragged around.
For example, users can animate a 3D model of an animal by dragging its feet forward.
This problem is severely ill-posed and typically under-constrained, as there are many possible deformations that can be matched with the provided partial surface deformations of handles, especially for large surface deformations.
Thus, strong priors encoding deformation regularity are necessary to tackle this problem.
Physics and differential geometry provide solutions that use various analytical priors which define natural-looking mesh deformations, such as elasticity~\cite{terzopoulos1987elastically,alexa2000rigid}, Laplacian smoothness~\cite{lipman2004differential, sorkine2004laplacian, zhou2005large}, and rigidity~\cite{sorkine2007rigid,sumner2007embedded,levi2014smooth} priors.
They update mesh vertex coordinates by iteratively optimizing energy functions that satisfy constraints from both the pre-defined deformation priors and given handle locations.
Although these algorithms can preserve geometric details of the original source model, they still have limited capacity to model realistic deformations, since the deformation priors are region independent, e.g., the head region deforms in a similar way as the tail of an animal, resulting in unrealistic deformation states.
Hence, motivated by the recent success of deep neural networks for 3D shape modeling~\cite{mescheder2019occupancy,park2019deepsdf,chen2019learning,xu2019disn,tang2019skeleton,chibane2020implicit,peng2020convolutional,jiang2020local,atzmon2020sal,gropp2020implicit,chabra2020deep,tretschk2020patchnets,tang2021sa,chen2021model}, we propose to learn shape deformation priors of a specific object class, e.g., quadruped animals, to complete surface deformations beyond observed handles.
We formulate the following properties of such a learned model; (1) it should be robust to different mesh quality and number of vertices, (2) the source mesh is not limited to canonical pose (i.e., the input mesh can have arbitrary pose), and (3) it should generalize well to new deformations.
Towards these goals, we represent deformations as a continuous deformation field which is defined in the near-surface region to describe the space deformation caused by the corresponding surface deformation.
The continuity property enables us to manipulate meshes with infinite number of vertices and disconnected components. 
%
To handle source meshes in arbitrary poses, we learn shape deformations via canonicalization.
Specifically, the overall deformation process consists of two stages: arbitrary-to-canonical transformation and canonical-to-arbitrary transformation.
To obtain more detailed surface deformations and better generalization capabilities to unseen deformations, we propose to learn local deformation fields conditioned on local latent codes encoding geometry-dependent deformation priors, instead of global deformation fields conditioned on a single latent code.
To this end, we propose Transformer-based Deformation Networks (TD-Nets), which learns encoder-based local deformation fields  on point cloud approximations of the input mesh.
Concretely, TD-Nets encode an input point cloud with surface geometry information and incomplete deformation flow into a sparse set of local latent codes and a global feature vector by using the vector attention blocks proposed in ~\cite{zhao2021point}.
The deformation vectors of spatial points are estimated by an attentive decoder, which aggregates the information of neighboring local latent codes of a spatial point based on the feature similarity relationships.
The aggregated feature vectors are finally passed to a multi-layer-perceptron (MLP) to predict displacement vectors which can be applied to the source mesh to compute the final output mesh.
%
%

\medskip
\noindent
To summarize, we introduce transformer-based local deformation field networks which are capable to learn shape deformation priors for the task of user-driven shape manipulation.
The deformation networks learn a set of anchor features based on a vector attention mechanism, enhancing the global deformation context, and selecting the most informative local deformation descriptors for displacement vector estimations, leading to an improved generalization ability to new deformations.
In comparison to classical hand-crafted deformation priors as well as recent neural network-based deformation predictors, our method achieves more accurate and natural shape deformations.
\section{Related Work}
User-guided shape manipulation lies at the intersection of computer graphics and computer vision.
Our proposed method is related to polygonal mesh geometry processing, neural field representations, as well as vision transformers.

\paragraph{Optimization-based Shape Manipulation.}
Classical methods formulate shape manipulation as a mathematical optimization problem.
They perform mesh deformations by either deforming the vertices~\cite{botsch2007linear, sorkine2006differential} or the 3D space~\cite{jacobson2011bounded, bechmann1994space, levi2014smooth, milliron2002framework, sederberg1986free}.
Performing mesh deformation without any other information about the target shape, but only using limited user-provided correspondences is an under-constrained problem. 
%
To this end, the optimization methods require deformation priors to constraint the deformation regularity as well as the smoothness of the deformed surface.
Various analytic priors have been proposed which encourage smooth surface deformations, such as elasticity~\cite{terzopoulos1987elastically,alexa2000rigid}, Laplacian smoothness~\cite{lipman2004differential, sorkine2004laplacian, zhou2005large}, and rigidity~\cite{sorkine2007rigid,sumner2007embedded,levi2014smooth}.
These methods use efficient linear solvers to iteratively optimize energy functions that satisfy constraints from both the pre-defined deformation prior and provided handle movements.
Recently, NFGP~\cite{yang2021geometry} was proposed to optimize neural networks with non-linear deformation regularizations.
Specifically, it performs shape deformations by warping the neural implicit fields of the source model through a deformation vector field, which is constrained by modeling implicitly represented surfaces as elastic shells.
\rev{NeuralMLS~\cite{shechter2022neuralmls} learned a geometry-aware weight function of a shape and given control points for moving least squares(MLS) deformations, which smoothly interpolates the control point displacements over space.}
Although they can preserve many geometric details of the source shape, they struggle to model complex deformations, as local surfaces are simply constrained to be transformed in a similar manner.
In contrast, we aim to learn deformation priors based on local geometries to infer hidden surface deformations.
%

\paragraph{Learning-based Shape Reconstruction and Manipulation.}
%
Learning-based shape manipulation has been studied to learn  shape priors based on shape auto-encoding or auto-decoding.
\cite{zheng2021deep, deng2021deformed, hao2020dualsdf, jiang2020shapeflow} map a class of shapes into a latent space.
During inference, given handle positions as input, they find an optimal latent code whose 3D interpretation is the most similar to the observation.
In contrast, we learn explicit deformation priors to directly predict 3D surface deformations.
Jakab et al.~\cite{jakab2021keypointdeformer} proposed to control shapes via unsupervised 3D keypoint discovery.
Instead, we use partial surface deformations represented by handle displacements as input observations, rather than keypoint displacements.
There exist a series of methods that use deep neural networks to complete non-rigid shapes~\cite{jiang2020shapeflow, palafox2021npms, bozic2021neural, li20214dcomplete, tang2021learning, saito2021scanimate, wang2021metaavatar, burov2021dynamic} from partial scans.
Our task is partially related to this task, but our shape manipulation task from user input requires completion of the deformation field.
In contrast to shape completion, our setting is more under-constrained, as the user-provided handle correspondences are very sparse and more incomplete than partial point clouds from scans.
Recent methods for clothed-human body reconstruction choose to canonicalize the captured scan into a pre-defined T-pose~\cite{wang2021locally, mihajlovic2021leap, chen2021snarf} using the skeletal deformation model of SMPL~\cite{smpl} or STAR~\cite{star} which can also be used to later animate the human.
Inspired by this, we also perform a canonicalization to enable editing of source meshes with arbitrary poses, before applying the actual deformation towards the target pose handles.
%


\paragraph{Continuous Neural Fields.}
Continuous neural field representations have been widely used in 3D shape modeling~\cite{mescheder2019occupancy, chen2019learning, park2019deepsdf} and 4D dynamics capture~\cite{niemeyer2019occupancy, tang2021learning, bozic2021neural, palafox2021npms, li20214dcomplete}.
Recent work that represents 3D shapes as continuous signed distance fields~\cite{atzmon2020sal, xu2019disn, gropp2020implicit, chabra2020deep, tretschk2020patchnets} or occupancy fields~\cite{mescheder2019occupancy, chen2019learning,chibane2020implicit,mi2020ssrnet,peng2020convolutional,jiang2020local,tang2021skeletonnet,tang2021sa,giebenhain2021air,zhang2021training}  can theoretically obtain volumetric reconstructions with infinite resolutions, as they are not bound to the resolution of a discrete grid structure.
Similarly, we learn continuous deformation fields defined in 3D space for shape deformations~\cite{tang2019skeleton,jiang2020shapeflow,yang2021geometry,hui2022neural}.
Due to the continuity of the deformation fields, our method is not limited by the number of mesh vertices, or disconnected components.
\rev{Different from ShapeFlow~\cite{jiang2020shapeflow}, OFlow~\cite{niemeyer2019occupancy}, LPDC-Net~\cite{tang2021learning} and NPMs~\cite{palafox2021npms} that learn a deformation field from a single latent code,
}
inspired by local implicit field learning~\cite{chibane2020implicit, peng2020convolutional, tang2021sa, giebenhain2021air, zhang20223dilg}, we model the deformation field as a composition of local deformation functions, improving the representation capability of describing complex deformations as well as  generalization to new deformations.

\paragraph{Visual Transformers.}
Recently, transformer architectures~\cite{vaswani2017attention} from natural language processing have revolutionized many computer vision tasks, including image classification~\cite{dosovitskiy2020image, wang2018non}, object recognition~\cite{carion2020end}, semantic segmentation~\cite{zheng2021rethinking}, or 3D reconstruction~\cite{bozic2021transformerfusion, yu2021cofinet, giebenhain2021air, zhang20223dilg, rao2022patchcomplete}.
We refer the reader to ~\cite{han2020survey} for a detailed survey of visual transformers.
In this work, we propose the usage of a transformer architecture to learn deformation fields.
Given the input point cloud sampled from the source mesh with partial deformation flow (defined by the user handles), we employ the vector attention blocks from Point Transformer~\cite{zhao2021point} as a main point cloud processing module to extract a sparse set of local latent codes, enhancing the global understanding of deformation behaviours.
Based on the obtained local deformation descriptors, our attentive deformation decoder learns to attend to the most informative features from near-by local codes to predict a deformation field.
\section{Approach}
\label{SecApproach}

\begin{figure}[!t]
    \centering
    \includegraphics[width=\linewidth]{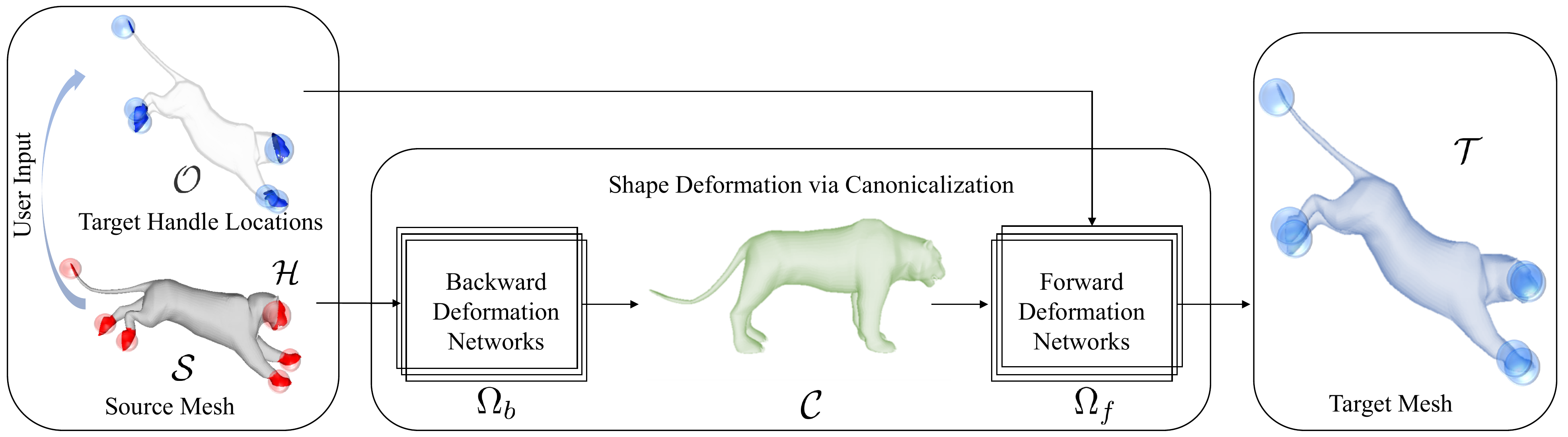}
    \caption{
        \textbf{Overview}. Given a source mesh $\mathcal{S}$ with sparse handles $\mathcal{H}$ (\red{red} circles) and their respective target locations $\mathcal{O}$ (\blue{blue} circles) as input, our method deforms the mesh to the target mesh $\mathcal{T}$ via canonicalization $\mathcal{C}$.
        The backward $\Omega_b$ and forward $\Omega_f$ deformation networks store the deformation priors that allow our method to produce consistent and natural-looking outputs. 
        }
    \label{fig:pipeline}
\end{figure}

Given a source mesh $\mathcal{S} =\{ \mathcal{V}, \mathcal{F} \}$ where $\mathcal{V}$ and $\mathcal{F}$ denote the set of vertices and the set of faces, respectively, we aim to deform $\mathcal{S}$ to obtain a target mesh $\mathcal{T}$ by selecting a sparse set of mesh vertices $\mathcal{H}=\{\mathbf{h}_i\}_{i=1}^{\ell}$ as handles, and dragging them to target locations $\mathcal{O}=\{\mathbf{o}_i\}_{i=1}^{\ell}$.
%
%
%
The key idea in this work is to use deformation priors to complete hidden surface deformations.
Specifically, the goal is to learn a continuous deformation field $\mathbf{D}$ defined in 3D space, from which we can obtain the deformed mesh $\mathcal{T}' =\{ \mathcal{V} + \mathbf{D}(\mathcal{V}), \mathcal{F} \}$ through vertex deformations of the source mesh $\mathcal{S}$.
The overall pipeline of the proposed approach is shown in \Cref{fig:pipeline}.
Our method can be applied to input meshes in arbitrary poses by leveraging learned shape deformation via canonicalization (see \Cref{SubSecCano}).
To represent the underlying deformation prior, we propose neural deformation fields as described in \Cref{SubSecTransDef} which can be learned from large deformation datasets (see \Cref{SubSecLoss}).

\subsection{Learning Shape Deformations via Canonicalization}
\label{SubSecCano}

\rev{To ensure robustness} w.r.t. varying input mesh quality (topology and resolution), we operate \rev{on point clouds instead} of meshes.
Specifically, we sample a point cloud $\mathcal{P}_\mathcal{S}=\{\mathbf{p}_i\}_{i=0}^n \in \mathbb{R}^{n \times 3}$ from $\mathcal{S}$ of size $n=5000$. 
We define the target handle point locations $\mathcal{P}_\mathcal{O}=\{\mathbf{o}_i\}_{i=0}^n  \in \mathbb{R}^{n \times 3}$, where we use zeros to represent unknown point flows.
Further, to avoid the ambiguity of zero point flow, we define the corresponding binary user handle masks $\mathcal{M} =\{b_i\}_{i=0}^n \in \mathbb{R}^{n}$ where $b_i=1$ if $\mathbf{p}_i$ is a handle or otherwise $b_i=0$.

To learn the shape transformation between two arbitrary non-rigidly deformed poses, one can learn deformation fields that directly map the source deformed space to target space.
However, it would be difficult to learn the deformation priors well, as there could be infinite deformation state transformation pairs.
To decrease the learning complexity, we introduce a canonical space as an intermediate state.
We divide the shape transformation process into two steps; a backward deformation that aligns the source deformed space to canonical space, and a forward deformation that maps the canonical space to the target deformation space.
Concretely, $\mathcal{P}_\mathcal{S}$ is passed into the backward transformation network $\Omega_b$ to learn the backward deformation field $\mathbf{D}_b$ which transforms the input shape $\mathcal{P}_\mathcal{S}$ into a canonical pose $\mathcal{P}_\mathcal{C}'$.
Similarly, the querying \rev{non-surface} point set $\mathcal{Q}_\mathcal{S}=\{\mathbf{q}_i\}_{i=0}^m\in \mathbb{R}^{m \times 3}, m=5000$ randomly sampled in the 3D space of $\mathcal{S}$ is also mapped to canonical space through  $\mathcal{Q}'_\mathcal{C} = \mathcal{Q}_\mathcal{S} + \mathbf{D}_b(\mathcal{Q}_\mathcal{S})$.
Lastly, given $\mathcal{P}_\mathcal{C}'$, $\mathcal{M}$, and $\mathcal{P}_\mathcal{O}$ as input, a forward transformation network $\Omega_f$ is learned to represent the forward deformation field $\mathbf{D}_f$ that predicts final locations $\mathcal{Q}'_\mathcal{T} = \mathcal{Q}'_\mathcal{C} + \mathbf{D}_f(\mathcal{Q}'_\mathcal{C})$. 

\subsection{Transformer-based Deformation Networks (TD-Nets)}
\label{SubSecTransDef}
\begin{figure}[!t]
    \centering
    \includegraphics[width=\linewidth]{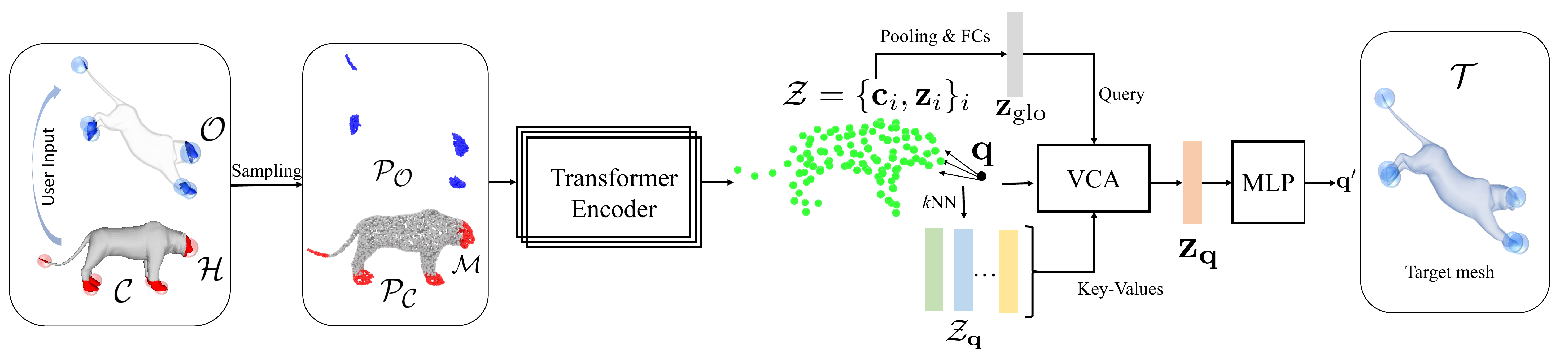}
    \caption{
    \textbf{Transformer-based Forward Deformation Networks}.
    Given a canonical mesh $\mathcal{C}$ with handle positions $\mathcal{H}$ (\red{red} circles) and desired handle locations $\mathcal{O}$ (\blue{blue} circles), we perform surface sampling to obtain a point cloud $\mathcal{P}_\mathcal{C}$ with additional channels of handle mask $\mathcal{M}$ and point flow $\mathcal{P}_\mathcal{O}$.
    A point-transformer encoder is devised to extract a sparse set of local latent codes $\mathcal{Z} = \{ \mathbf{c}_i,  \mathbf{z}_i\}_{i}$ from this point cloud, where $\mathbf{c}_i$ are the anchor positions of the latent features $\mathbf{z}_i$.
    For a specific point $\mathbf{q}$ in 3D space (i.e. a vertex from the source mesh), based on the $\mathbf{z}_{\glo}$, a vector cross attention ($\VCA$) block is used to effectively fuse the information of $\mathcal{Z}_\mathbf{q}$ into $\mathbf{z_q}$ from the $k$ nearest neighbouring latent codes of $\mathbf{q}$.
    Using a multi-layer perceptron (MLP) conditioned on $\mathbf{z_q}$, we predict the deformed location $\mathbf{q}'$ in the target space.
    %
    }
    \label{fig:transdeform}
\end{figure}

The deformation via canonicalization is based on two deformation field predictors (forward and backward deformations).
Both networks share the same architecture, thus, in the following, we will only describe the forward deformation network as visualized in \Cref{fig:transdeform} while the backward deformation network is analogous.
It consists of a transformer-based deformation encoder and a vector cross attention-based decoder network.

\paragraph{Point transformer encoder.}
Given a point set $\mathcal{P}_\mathcal{C}$ with handle locations $\mathcal{P}_\mathcal{O}$ and a binary mask $\mathcal{M}$ as inputs, we use point transformer layers from~\cite{zhao2021point} to build our encoder modules.
The point transformer layer is based on the vector attention mechanism~\cite{zhao2020exploring}.
Let $\mathcal{X} = \{\mathbf{x}_i , \mathbf{f}_i \}_i$  and $\mathcal{Y} =  \{\mathbf{y}_i , \mathbf{g}_i \}_i$ be the query and key-value sequences, where $\mathbf{x}_i$ and $\mathbf{y}_i$ denote the coordinates of query and key-value points with corresponding feature vectors $\mathbf{f}_i$ and $\mathbf{g}_i$.
%
The vector cross attention operator $\VCA$ is defined as:
\begin{equation}
    \VCA(\mathcal{X}, \mathcal{Y}): ~~ \mathbf{f}_i' = \sum_{j \in \mathcal{N}_i} \rho( \gamma( \varphi(\mathbf{g}_j) - \psi(\mathbf{f}_i) + \delta )) \odot (\alpha( \mathbf{f}_i) + \delta) ,
    \label{equa:VCA}
\end{equation}
where $\mathbf{f}_i'$ are the aggregated features, $\varphi$, $\psi$, and $\alpha$ are linear projections implemented by a fully-connected layer.
$\gamma$ is a mapping function implemented by a two-layer MLP to predict attention vectors.
$\rho$ is the attention weight normalization function, in our case \emph{softmax}. $\delta := \theta (\mathbf{x}_i - \mathbf{y}_j)$ is the positional embedding module~\cite{vaswani2017attention, mildenhall2020nerf} implemented by a two linear layers with a single ReLU~\cite{nair2010rectified}.
It leverages relatively positional information of $\mathbf{x}_i$ and $\mathbf{y}_j$ to benefit the network training.
Then, with the definition of VCA, the vector self-attention operator $\VSA$ can be defined as: 
\begin{equation}
    \VSA(\mathcal{X}): = \VCA(\mathcal{X}, \mathcal{X}).
    \label{equa:VSA}
\end{equation}
Based on VCA and VSA, we can define two basic modules to build our encoder network, i.e. the \emph{point transformer block} (PTB) and the \emph{point abstraction block} (PAB).
The definition of the point transformer block $\PTB$ is a combination of the BatchNorm (BN) layer~\cite{ioffe2015batch}, VSA, and residual connections, formulated as:
\begin{equation}
     \PTB(\mathcal{X}): = \BN (\mathcal{X} + \VSA(\mathcal{X})).
    \label{equa:PTB}
\end{equation}
For each point $\mathcal{X}_i$, it encapsulates the information from $k_{\enc}=16$ nearest neighborhoods while keeping the point's position $\mathbf{x}_i$ unchanged.
The point abstraction block $\PAB$ consists of farthest point sampling (FPS), BN, VCA, and VSA, which is defined as follow:
\begin{equation}
     \PAB(\mathcal{X}): = \BN(\FPS(\mathcal{X}) + \VSA(\VCA(\FPS(\mathcal{X}), \mathcal{X})).
    \label{equa:PTB}
\end{equation}
%

%
The point cloud $\mathcal{P}_\mathcal{C} $ with handle mask $\mathcal{M}$ and flow $\mathcal{P}_\mathcal{O}$ as additional channels are passed to a point transformer block (PTB) to obtain a feature point cloud
$\mathcal{Z}_0 = \{ \mathbf{c}_i^0,  \mathbf{z}_i^0\}_{i=1}^{n}$.
By using two consecutive point abstraction blocks (PABs) with intermediate set size of $n_1=500$ and $n_2=100$, we obtain $\mathcal{Z}_1 = \{ \mathbf{c}_i^1,  \mathbf{z}_i^1\}_{i=1}^{n_1}$ and $\mathcal{Z}_2 = \{ \mathbf{c}_i^2,  \mathbf{z}_i^2\}_{i=1}^{n_2}$.
To enhance global deformation priors, we stack 4 point transformer blocks with full self-attention whose $k_{\enc}$ is set to 100 to exchange the global information in the whole set of $\mathcal{Z}_2$.
By doing so, we can obtain a sparse set of local deformation descriptors $\mathcal{Z} = \{ \mathbf{c}_i,  \mathbf{z}_i\}_{i=1}^{100}$ that are anchored in $\{ \mathbf{c}_i \}$.
Finally, we perform a global max-pooling operation followed by two linear layers to obtain the global latent vector $\mathbf{z}_{\glo}$.

\noindent \textbf{Attentive deformation decoder.}
Based on the learned local latent codes $\mathcal{Z} = \{ \mathbf{c}_i,  \mathbf{z}_i\}_{i=1}^{100}$ and global latent vector $\mathbf{z}_{\glo}$, the deformation decoder defines the forward deformation function $\mathbf{D}_f: \mathbb{R}^3 \xrightarrow{} \mathbb{R}^3$, which maps a point $\mathbf{q}$ from the canonical space of $\mathcal{C}$ to the 3D space of $\mathcal{T}$.
Similar to tri-linear interpolation operations in grid-based implicit field learning, a straightforward way to find the corresponding feature vector $\mathbf{z_q}$ is to use the weighted combination of $k_{\dec}=16$ nearby local codes $\mathcal{Z}_\mathbf{q} = \{\mathbf{c}_k, \mathbf{z}_k\}_{k=1}^{k_{\dec}}$.
Intuitively, the weight is inversely proportional to the euclidean distance between $\mathbf{q}$ and the anchoring location $\mathbf{c}_k$~\cite{peng2020convolutional}.
However, distance-based feature queries ignore the relationships between deformation descriptors.
Thus, we propose to obtain $\mathbf{z_q}$ by adaptively aggregating information of $\mathcal{Z}_\mathbf{q}$ based on the vector cross-attention operator:
\begin{equation}
    \mathbf{z_q} = \VCA(\{\mathbf{q}, \mathbf{z}_{\glo}\}, \mathcal{Z}_\mathbf{q}).
    \label{equa:Dec}
\end{equation}
The local information aggregation enables us to flexibly search the local deformation priors, thus, improving the generalizability to new deformations.
Finally, the $\mathbf{z_q}$ is fed into an MLP composed of five Res-FC blocks to estimate the associate location $\mathbf{q}'= \mathbf{q} + \mathbf{D}_f(\mathbf{q};\mathbf{z_q})$ in the target space.

\subsection{Training Objectives}
\label{SubSecLoss}
For training, we need a set of triplets $(\mathcal{S}, \mathcal{C}, \mathcal{T})$ with dense correspondences, from which we can randomly sample surface point clouds $(\mathcal{P}_\mathcal{S}, \mathcal{P}_\mathcal{C}, \mathcal{P}_\mathcal{T})$ of size $n$ and querying \rev{non-surface} points $(\mathcal{Q}_\mathcal{S}, \mathcal{Q}_\mathcal{C}, \mathcal{Q}_\mathcal{T})$ of size $m$ in the 3D space.
To optimize the backward deformation networks, we employ the mean $\ell_2$ distance error that measures the difference between deformed points from source space and their ground-truths in the canonical space:
\begin{equation}
    L_b = ||\Omega_b(\mathcal{P}_\mathcal{S}) - \mathcal{P}_\mathcal{C}||_2^2 +  ||\Omega_b(\mathcal{Q}_\mathcal{S}) - \mathcal{Q}_\mathcal{C}||_2^2 .
    \label{equa:Lback}
\end{equation}
Similarly, to optimize the forward deformation networks, we use the following loss function:
\begin{equation}
    L_f = ||\Omega_f(\mathcal{P}_\mathcal{C}) - \mathcal{P}_\mathcal{T}||_2^2 +  ||\Omega_b(\mathcal{Q}_\mathcal{C}) - \mathcal{Q}_\mathcal{T}||_2^2 
    \label{equa:Lfor}
\end{equation}
The total loss function for source-target shape deformations is defined as:
\begin{equation}
     L_{\text{total}} = ||\Omega_f(\Omega_b(\mathcal{P}_\mathcal{S})) - \mathcal{P}_\mathcal{T}||_2^2 +  ||\Omega_f(\Omega_b(\mathcal{Q}_\mathcal{S})) - \mathcal{Q}_\mathcal{T}||_2^2 .
    \label{equa:Ltotal}
\end{equation}

%
\section{Experiments}
\label{SecExp}
\paragraph{Dataset.}
Our experiments are performed on the DeformingThing4D-Animals~\cite{li20214dcomplete} dataset which contains 1494 non-rigidly deforming animations with various motions comprising 40 identities of 24 categories.
For the train/test split, we divide all animations into training (1296) and test (198).
Similar to the D-FAUST~\cite{bogo2017dynamic} used in OFlow~\cite{niemeyer2019occupancy}, the test set is composed of two subsets:
(S1) contains 143 sequences of new motions for seen train identities,
and (S2) contains 55 sequences of unseen individuals (and thus also new motions).
During training, we randomly sample two frames from an identity as source-target deformation pairs.
During inference, we consider the first frame of an animation as source mesh, and other frames as target meshes.
%
To evaluate the generalization ability to unseen identities, we evaluate the pre-trained models on the animal dataset used in Deformation Transfer~\cite{sumner2004deformation}. 
For the quantitative comparison on each test subset, we compute evaluation metrics for 300 randomly sampled pairs.
%
In addition, we also include comparisons on another animal dataset used in TOSCA~\cite{rodola2017partial}.
TOSCA~\cite{rodola2017partial} does not have correspondences between different poses of the same animal, and hence does not easily provide handle displacements as input. Thus, we provide a qualitative comparison under the setting of using user-specified handles as inputs.
%
\paragraph{Implementation details.}
Our approach is built on the PyTorch library~\cite{paszke2019pytorch}.
Please refer to the supplementary material for the details of our network architecture.
%
Our model consists of two training stages.
We use an Adam~\cite{kingma2014adam} optimizer with $\beta_1=0.9$, $\beta_2= 0.999$, and $\epsilon=10^{-8}$.
In the first stage, we train the forward and backward deformation networks individually.
Specifically, the backward and forward deformation networks are respectively optimized by the objective described in Equations~\ref{equa:Lback} or~\ref{equa:Lfor} using a batch size of 16 with the learning rate of 5e-4 for 100 epochs.
In the second stage, the whole model is trained according to Equation~\ref{equa:Ltotal} in an end-to-end manner using a batch size of 6 with a learning rate of 5e-5 for 20 epochs. 

\paragraph{Baselines.}
We conduct comparisons against classical optimization-based and recent neural network-based methods.
For the former, we select a representative work, ARAP~\cite{sorkine2007rigid}, that constrains each local surface to be rigidly transformed as much as possible.
%
For the latter, we compare our method with the learning-based deformation predictor ShapeFlow~\cite{jiang2020shapeflow} that embeds each shape into a latent space and learns flow-based deformations among 3D shapes.
We also compare to NFGP~\cite{yang2021geometry}, a deep optimization method, which constrains the implicitly represented surfaces as elastic shells during the deformation process.
%
    \begin{figure}[h]
        \centering
        \includegraphics[width=\linewidth]{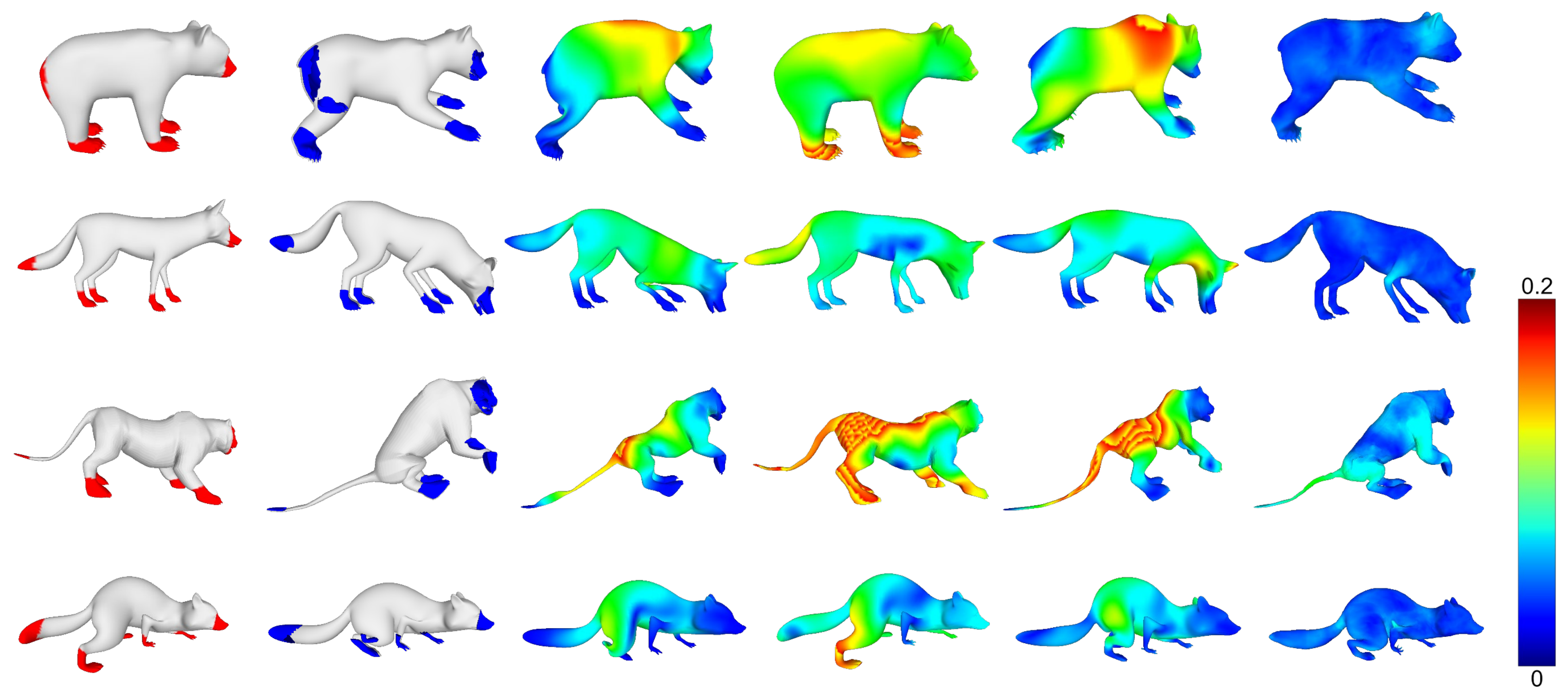}
             \begin{tabular}{p{60pt}p{55pt}p{50pt}p{60pt}p{55pt}p{50pt}}
                  Source mesh  & 
                  Target mesh &
                  ARAP~\cite{sorkine2007rigid} & 
                  ShapeFlow~\cite{jiang2020shapeflow} &
                  NFGP~\cite{yang2021geometry} & 
                  Ours \\ and handles & and handles & &
                \end{tabular}
        \caption{Comparison against ARAP~\cite{sorkine2007rigid}, ShapeFlow~\cite{jiang2020shapeflow}, and NFGP~\cite{yang2021geometry} on new motions. We visualize the vertex euclidean distance errors as color maps. 
        }
        %
        \label{fig:unseen_motion}
    \end{figure}
    
    \begin{figure}[h]
        \centering
        \includegraphics[width=\linewidth]{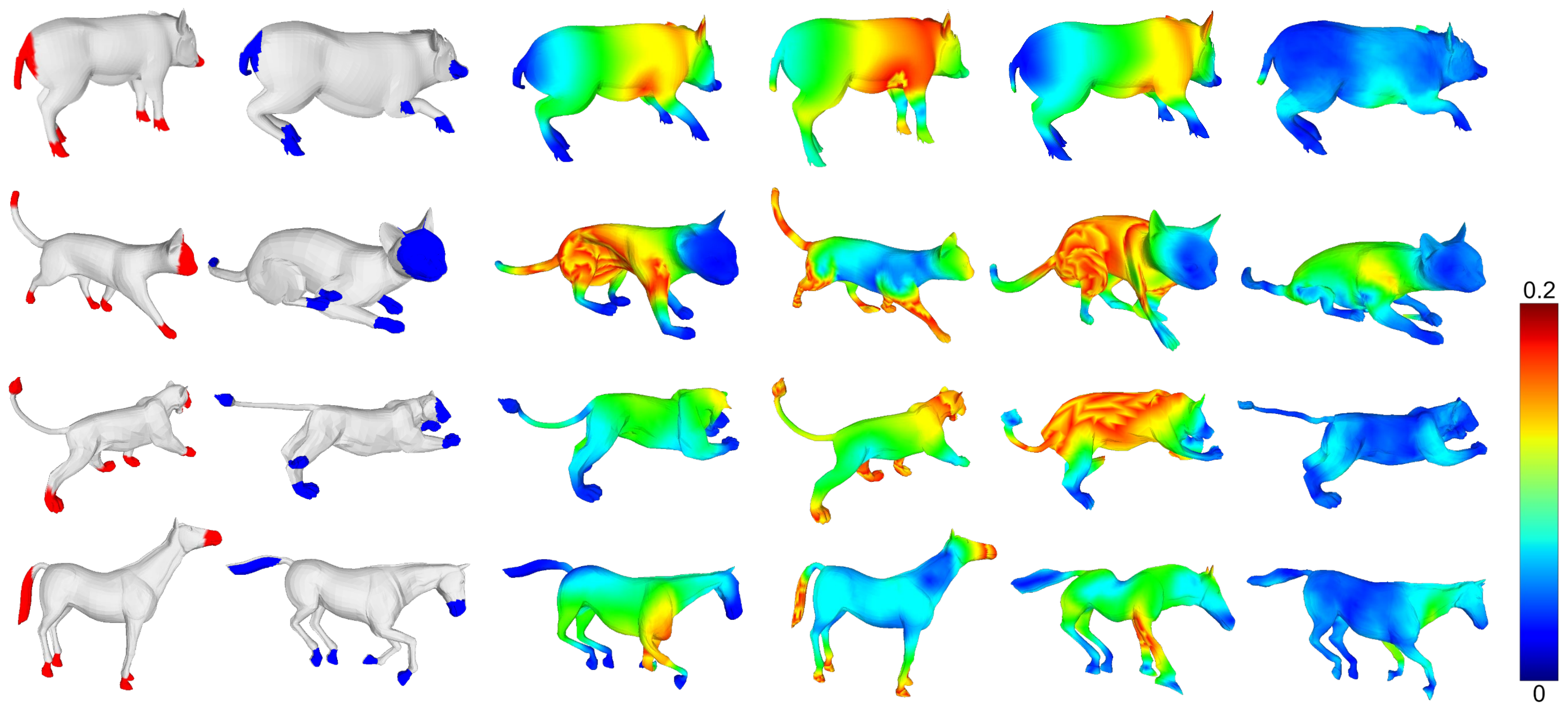}
        \begin{tabular}{p{60pt}p{55pt}p{50pt}p{60pt}p{55pt}p{50pt}}
                  Source mesh & 
                  Target mesh &
                  ARAP~\cite{sorkine2007rigid} & 
                  ShapeFlow~\cite{jiang2020shapeflow} &
                  NFGP~\cite{yang2021geometry} & 
                  Ours \\ and handles & and handles & &
        \end{tabular}
        \caption{Comparison against ARAP~\cite{sorkine2007rigid}, ShapeFlow~\cite{jiang2020shapeflow}, and NFGP~\cite{yang2021geometry} on the S2 test set of DeformingThing4D-Animals and unseen shapes of Deformation Transfer~\cite{sumner2004deformation}. We visualize the vertex euclidean distance errors as color maps.
        Our approach generalizes better in comparison to ShapeFlow and NFGP and produces natural looking deformations (in comparison, ARAP generates rubber-like deformations).} 
        \label{fig:unseen_identity}
    \end{figure}
    
\paragraph{Evaluation metrics.}
We consider $\ell_2$ distance error of mesh vertices ($\ell_2$ $\times 0.001$), Chamfer Distance (CD $\times 0.01$) of sampled point clouds of 30k points, and Face Normal Consistency (FNC $\times 0.01$) as primary evaluation metrics.
Please refer to the supplementary material for a detailed explanation of these metrics.
Note that for $\ell_2$ and CD, lower is better, while for FNC, higher is better.
%

\subsection{Comparisons}

\setlength{\tabcolsep}{4pt}
\begin{table*}[h]
	\renewcommand\arraystretch{1.2}
	\begin{center}
		\begin{tabular}{*{10}{c}}
			\toprule
			\multirow{2}*{Method} & \multicolumn{3}{c}{New motions (S1)}   & \multicolumn{3}{c}{Unseen identities (S2)}  & \multicolumn{3}{c}{Deformation Transfer} \\
			\cmidrule(lr){2-4} \cmidrule(lr){5-7}  \cmidrule(lr){8-10}
			 & $\ell_2$ $\downarrow$ & CD $\downarrow$ & FNC $\uparrow$   & $\ell_2$ $\downarrow$ & CD $\downarrow$ & FNC $\uparrow$   & $\ell_2$ $\downarrow$ & CD $\downarrow$ & FNC $\uparrow$  \\
			\midrule
			\midrule
           ARAP \cite{sorkine2007rigid} 
                   & 5.568  & 2.312  &  95.35  
                   & 9.794  & 2.308  &  94.89 
                   & 5.145  & 3.475  &  91.21  \\
                   
          ShapeFlow \cite{jiang2020shapeflow}   
                   & 21.03  & 3.494  & 89.69
                   & 32.08  & 3.925  & 90.73
                   & 33.72  & 4.093  & 86.36 \\
                      
           NFGP \cite{yang2021geometry}  
                   & 11.77  & 3.130  & 93.34
                   & 15.96  & 3.364  & 91.80  
                   & 18.90  & 4.150  & 82.54  \\
            
            \midrule
            Ours-VDF
                   & 3.590 & 1.887  &  86.01 
                   & 2.368 & 1.837  &  86.99
                   & 3.111 & 9.164  &  78.63 \\
        
            Ours-global       
                   & 2.970  & 1.546  &  93.30 
                   & 2.973  & 1.579  &  94.75  
                   & 2.636  & 8.453  &  84.59 \\
                   
            Ours-3D UNet
                   & 1.011 & 1.111 & 96.02
                   & 1.253 & 1.426 & 96.20 
                   & 4.553 & 2.362 & 88.31 \\
         
            Ours-PointNet++.
                   & 0.886 & 1.055 & 95.47 
                   & 1.231 & 1.364 & 95.37 
                   & 4.898 & 2.564 & 85.87 \\
            
            Ours-w/o atten dec.
                   & 1.184 & 1.210 & 95.64
                   & 1.227 & 1.417 & 96.16 
                   & 5.252 & 2.772 & 84.95 \\
                   
            Ours-w/o cano.
                   & 1.018 & 1.063 & 96.40
                   & 0.969 & 1.258 & 96.62
                   & 2.660 & 1.934 & 90.96 \\
                   
            Ours-full 
                   & \textbf{0.752} & \textbf{0.948}  & \textbf{96.59} 
                   & \textbf{0.795} & \textbf{1.241}  & \textbf{96.68} 
                   & \textbf{2.495} & \textbf{1.877}  & \textbf{91.40} \\

        \bottomrule
        \end{tabular}
        \caption{
        \rev{
            Quantitative comparisons on the S1 and S2 test sets of DeformingThing4D~\cite{li20214dcomplete} and the unseen identities of used in Deformation Transfer~\cite{sumner2004deformation}.
            }
        }
        \label{tab:comparison_all}
        \end{center}
        \vspace{-3mm}
\end{table*}

\begin{figure}[h]
    \centering
    \vspace{-3mm}
    \begin{tabular}{cl}
              \multicolumn{1}{c}{} & \multirow{25}{*}{\includegraphics[width=0.75\linewidth]{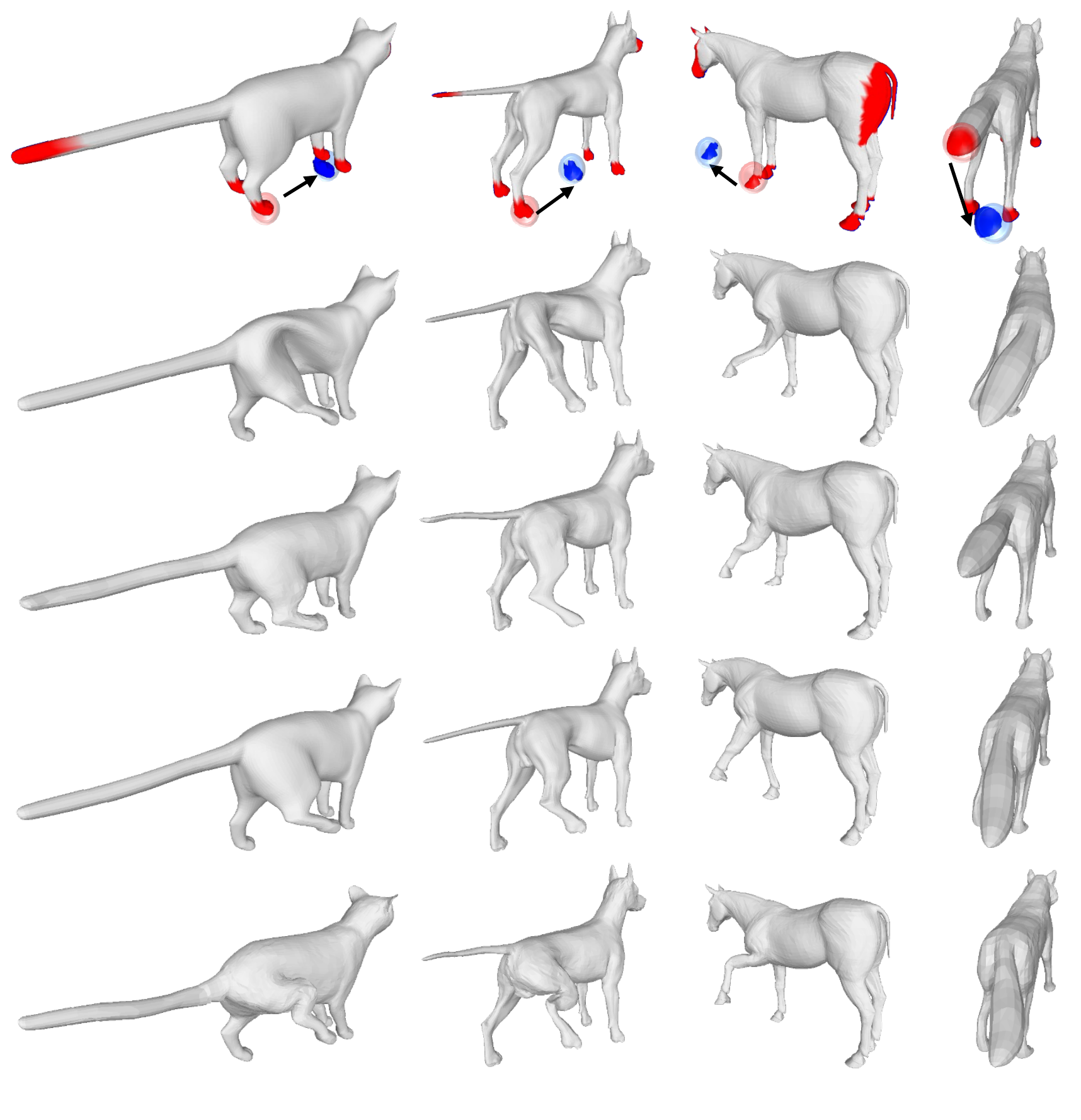}} \\
              \multicolumn{1}{c}{} & \\
               \multicolumn{1}{c}{} & \\
              \multicolumn{1}{c}{Source mesh,} & \\
              \multicolumn{1}{c}{handles and} & \\
              \multicolumn{1}{c}{target handles} & \\
               \multicolumn{1}{c}{} & \\
               \multicolumn{1}{c}{} & \\
               \multicolumn{1}{c}{} & \\
               \multicolumn{1}{c}{ARAP~\cite{sorkine2007rigid}} & \\
               \multicolumn{1}{c}{} & \\
               \multicolumn{1}{c}{} & \\
               \multicolumn{1}{c}{} & \\
               \multicolumn{1}{c}{} & \\
               \multicolumn{1}{c}{ShapeFlow~\cite{jiang2020shapeflow}} & \\
               \multicolumn{1}{c}{} & \\
               \multicolumn{1}{c}{} & \\
               \multicolumn{1}{c}{} & \\
               \multicolumn{1}{c}{} & \\
               \multicolumn{1}{c}{NFGP~\cite{yang2021geometry}} & \\
               \multicolumn{1}{c}{} & \\
                \multicolumn{1}{c}{} & \\
               \multicolumn{1}{c}{} & \\
               \multicolumn{1}{c}{} & \\
               \multicolumn{1}{c}{Ours} & \\
               \multicolumn{1}{c}{} & \\
               \multicolumn{1}{c}{} &
    \end{tabular}
    \caption{
    \rev{
        Comparison against ARAP~\cite{sorkine2007rigid}, ShapeFlow~\cite{jiang2020shapeflow} and NFGP~\cite{yang2021geometry} under the setting of user-specified handles on TOSCA dataset~\cite{rodola2017partial}. Our method visibly produces the best results.
        }
    }
    \label{fig:user_handles_tosca} 
\end{figure}

For a qualitative comparison, we visualize the vertex $\ell_1$ distance error maps of deformed meshes in \Cref{fig:unseen_motion} and \Cref{fig:unseen_identity}.
As can be seen, our method has lower vertex errors in the hidden surface regions since we use data-driven deformation priors, instead of employing hand-crafted regularizers to enforce surface smoothness.
The generalization ability to unseen deformations is improved by learning deformation fields for local surfaces, instead of modeling global deformations.
Compared to ARAP, ShapeFlow, and NFGP,  we can produce more realistic results for complicated actions in the 3rd and 4th rows of \Cref{fig:unseen_motion}.
The deformation results presented in \Cref{fig:unseen_identity} demonstrate that our method can generalize to unseen identities, and is also verified quantitatively in \Cref{tab:comparison_all}, where our method consistently outperforms all baselines.
\paragraph{User-specified handles.}
To evaluate the generalization performance of our approach on unseen identities using user-provided handle displacements that are used in interactive editing applications, we use random translations of handles applied to animals from TOSCA~\cite{rodola2017partial} as input. 
As depicted in \Cref{fig:user_handles_tosca}, our approach is able to produce naturally-looking deformation results, and shows its advantages compared to ARAP, ShapeFlow, and NFGP.
%
\rev{Note that for this demonstration of user-specified handles there exists no corresponding ground-truth.}
\subsection{Ablation studies}
\begin{figure}
        \centering
        \includegraphics[width=\linewidth]{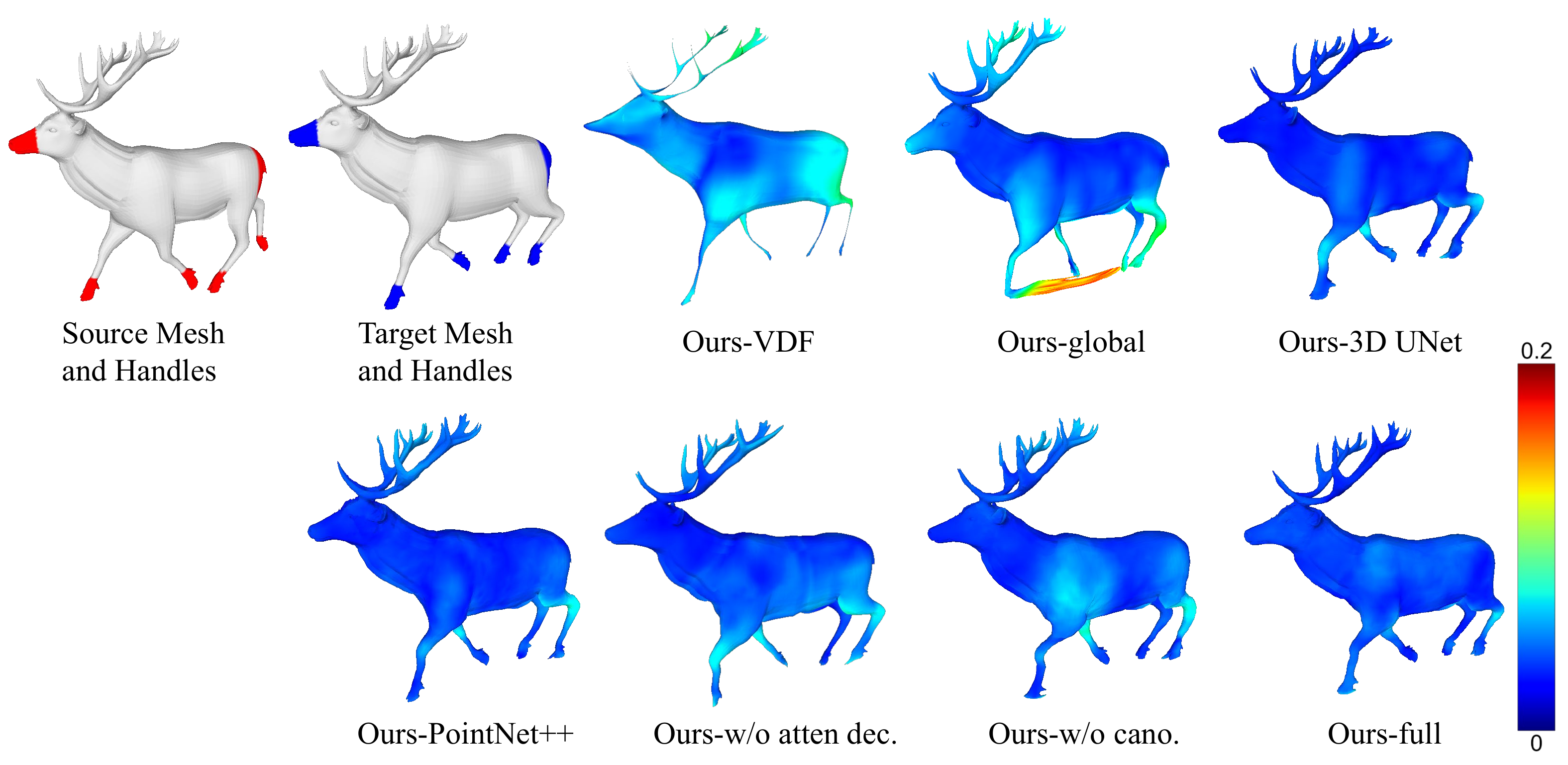}
        \caption{Qualitative ablation studies. Each component of our approach contributes to the final result that has the lowest reconstruction error.}
        \label{fig:ablation}
\end{figure}

To verify our final model choice, we conducted a series of ablation studies, where we analysed several variants of our deformation fields (see \Cref{tab:comparison_all} and \Cref{fig:ablation}).

\paragraph{Volumetric grids vs continuous fields.}
As continuous fields are not bound to the resolution of a discrete grid structure, it can  better represent  complex deformations.
The performance degrades when we learn grid-based volumetric deformation fields.
This can be seen in the experiment ``Ours-VDF" which uses a 3D U-Net~\cite{ronneberger2015u} to generate volumetric deformation fields of a fixed resolution $64^3$.

\paragraph{Global vs local deformation fields.}
``Ours-global" learns a global continuous field only conditioned on the global latent code.
This variant tends to lose detailed information about local surface deformations, and is more difficult to generalize to new motions or identities, leading to inferior results in comparison to our local deformation fields.

\paragraph{Network architectures (3D U-Net vs PointNet++ vs Point Transformer).}
Compared to grid-based and point-based local deformation descriptors learning, the point transformer-based encoder captures strong global contexts that enforce more global consistency constraints.
This provides performance improvements on surface accuracy of deformed meshes. 
To verify this, we conducted an experiment with ``Ours-3D-UNet," which learns a volumetric feature map through a 3D U-Net, and then predicts deformation fields based on queried features via tri-linear interpolation operations.
Additionally, we compare with ``Ours-PointNet++," which replaces the point transformer encoder with PointNet++~\cite{qi2017pointnet++}.

\paragraph{With vs without Attention-based feature querying.}
The attention-based feature query mechanism can flexibly and effectively select the most relevant deformation descriptors for a query point, resulting in improved performance over feature interpolation purely based on euclidean distances.
A deformation decoder that for example uses an interpolation with weights that are purely based on euclidean distance instead (``Ours-w/o atten. dec."), leading to significantly higher errors, particularly in terms of the $\ell_2$ vertex error.

\paragraph{With vs without canonical poses.} 
Learning shape deformations via canonicalization improves the generalization to source meshes in different poses.
Learning  without canonicalization ("Ours-w/o cano."), i.e., learning shape deformations directly between two arbitrary poses, results in considerably higher surface errors.
%
%
\subsection{Intermediate results of canonicalization}

    \begin{figure}
        \centering
        \includegraphics[width=\linewidth]{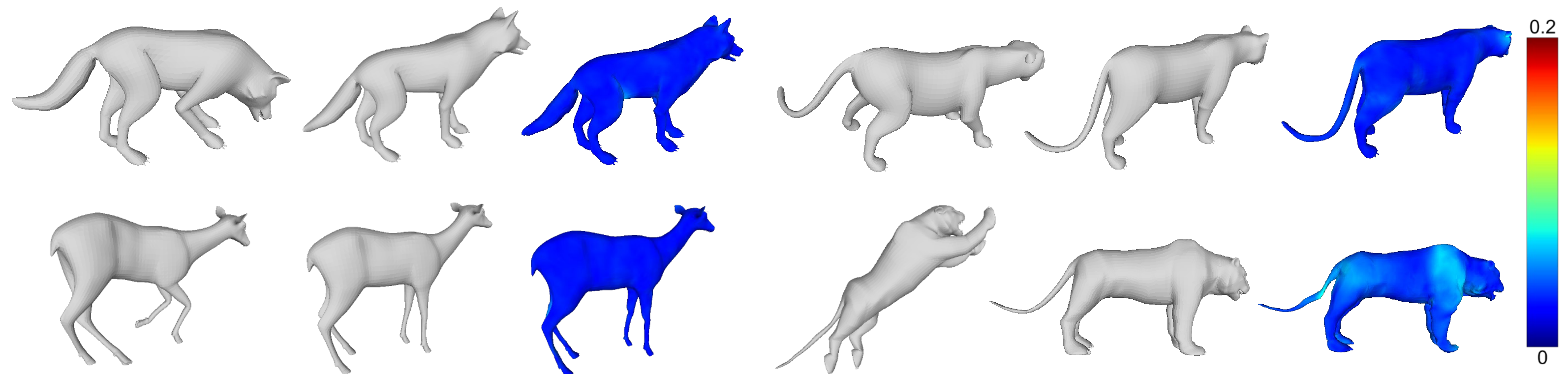}
         \begin{tabular}{@{\hspace*{3.5em}}p{50pt}p{50pt}p{68pt}p{45pt}p{55pt}p{55pt}}
                  (a) & 
                  (b) &
                  (c) &
                  (a) & 
                  (b) &
                  (c) 
        \end{tabular}
        \caption{The intermediate results of our canonicalization.
        (a) Source mesh. (b) Canonical mesh. (c) Our canonicalized mesh. }
        \label{fig:cano}
    \end{figure}

In~\Cref{fig:cano}, we visualize our intermediate results of canonicalization.
As can be seen, our method can project source meshes with arbitrary poses into a canonical space with a same pose.


\subsection{Limitations}
\label{SubSecLim}
While compelling results have been demonstrated for shape manipulation, a few limitations still exist in our approach that can be addressed in future work.
\rev{Our approach only needs sparse user input in form of handles which can be moved to create a new deformation state. While this allows for quick editing, a possible extension is to add rotations to the handles. This could be done by leveraging a different deformation representation such as a SE(3) field which is composed of a displacement and a rotation field. Note that our displacement representation is able to represent general deformations, but might require more user handles.}
%
%
%
%
Due to the limitations of the DeformingThing4D-Animals~\cite{li20214dcomplete} dataset in terms of available models and poses, our approach may suffer from the generalization to out-of-distribution models and extreme poses.
Additionally, the output of our model, as with other learning-based methods, may be affected by biases in the training dataset that can limit generalization.
We believe this issue can be relieved by a larger training dataset and a richer data augmentation strategy in future work.
Lastly,  our training scheme only considers handles that are selected from a set of candidate parts of the models, thus, limiting the regions the user can interact with.
Enriching the candidate handles during training is potentially helpful for allowing free handle placement.
%

\section{Conclusion}
\label{SecCon}
In this work, we introduced \textit{Neural Shape Deformation Priors}, a novel approach that learns mesh deformations of non-rigid objects from user-provided handles based on the underlying geometric properties of shapes.
To enable shape manipulation for source meshes with different poses, we choose to learn shape deformations via canonicalization where the source mesh is first transformed to the canonical space through a backward deformation field and then deformed to the target space through a forward deformation field.
For  deformation field learning, we propose Transformer-based Deformation Networks (TD-Net) that represent a shape deformation as a composition of local surface deformations.
Our experiments and ablation studies demonstrate that our method can be applied to challenging new deformations, outperforming classical optimization-based methods such as ARAP~\cite{sorkine2007rigid} and neural networks-based methods such as ShapeFlow~\cite{jiang2020shapeflow} and NFGP~\cite{yang2021geometry}, while showing a good generalization to previously unseen identities.
We see our method as an important step in the development of 3D modeling algorithms and softwares and hope to inspire more research in learning-based shape manipulation.

\paragraph{Societal impact.}

Our work provides an algorithm for natural-looking shape editing, which can simplify tedious procedures in 3D content creation and empower artists in the movie and game industries. It further has the potential to enrich 3D data with additional deformed shapes, and could thus help improve the performance of other practical application techniques that rely on large quantities of 3D ground-truth for training. Yet, misuse of our shape manipulation algorithm could enable fraud or offensive content generation.

\paragraph{Acknowledgement.}
This work is supported by a TUM-IAS Rudolf M{\"{o}}{\ss}bauer Fellowship, the ERC
Starting Grant Scan2CAD (804724), and Sony Semiconductor Solutions Corporation. We would also like to thank Angela Dai for the video voice over.




\bibliographystyle{splncs04}
\bibliography{ref}

\newpage
\appendix
\vspace{3mm}
\section*{\LARGE \centering Neural Shape Deformation Priors \\
      -- Supplementary Material --}
\vspace{9mm}
\maketitle
Our Neural Shape Deformation Priors method is based on transformer-based deformation networks that represent the deformation as a composition of local surface deformations.
The underlying architectures are discussed in \Cref{SecNet}. 
The used evaluation metrics are detailed in \Cref{SecEva}.
Our notations are further explained in \Cref{SecNotation}. And more details about data-preprocessing are given in \Cref{SecData}.
In addition to the results shown in the main paper, we conducted further experiments (see \ref{SecAdd}).
While our method exhibits good generalization to unseen poses and shapes, we discuss and show failure cases in \Cref{SecLim}.

\section{Network Architectures}
\label{SecNet}

\begin{figure}[!htb]
    \centering
    \includegraphics[width=\linewidth]{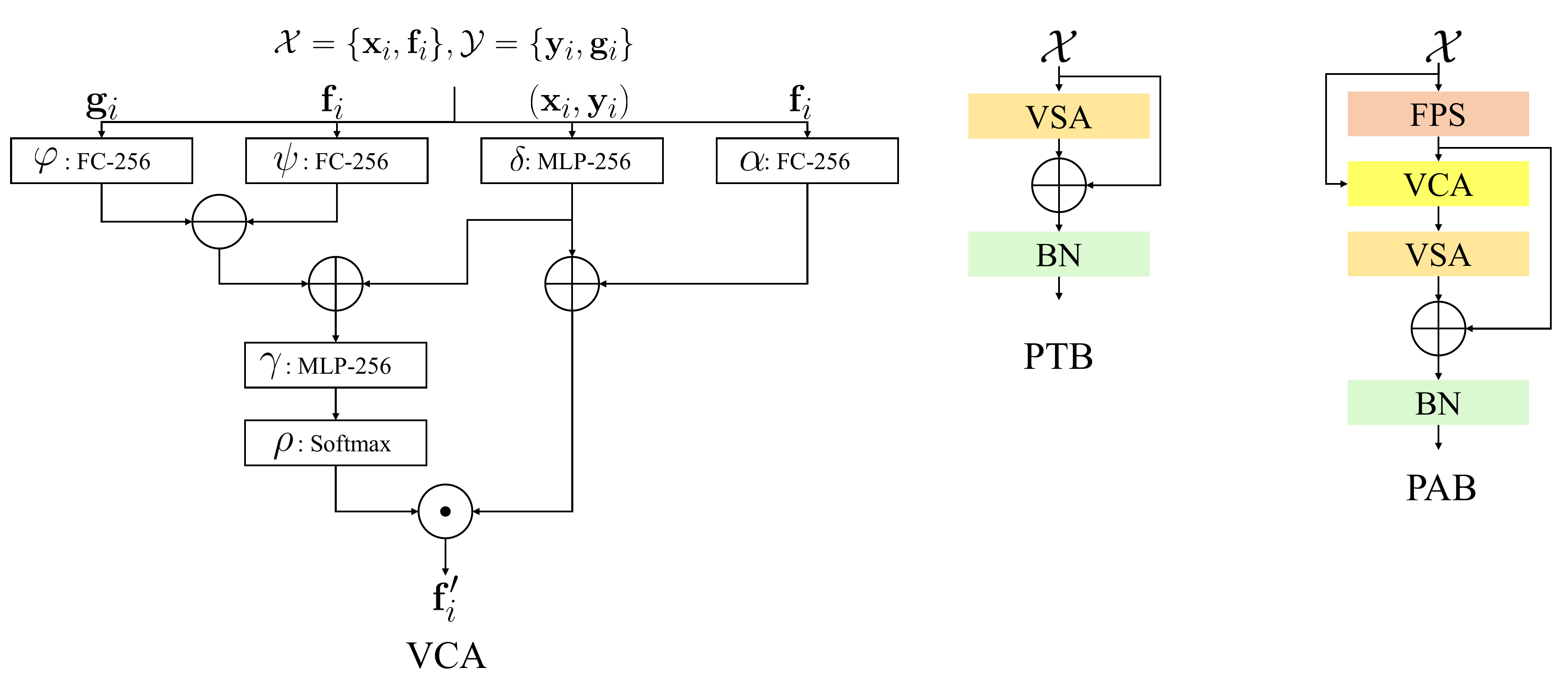}
    \caption{
    \textbf{Vector Cross Attention (VCA), Point Transformer Block (PTB), and Point Abstraction Block (PAB)}.
    }
    \label{fig:vca}
\end{figure}

\paragraph{Vector Cross Attention:}
In \Cref{fig:vca}, we illustrate the architecture of vector cross attention~\cite{zhao2020exploring} (VCA) which is a building block of our transformer-based deformation network (see Figure 3 in the main paper).
The feature vectors $\mathbf{g}_i$ and $\mathbf{f}_i$ are transformed with three linear projectors $\varphi(\mathbf{g}_i)$, $\psi(\mathbf{f}_i)$ and $\alpha(\mathbf{f}_i)$, each of which is a fully-connected layer.
To leverage relatively positional information of $\mathbf{f}_i$ and $\mathbf{g}_i$, 
 $\mathbf{x}_i - \mathbf{y}_i$ is encoded by a positional embedding module~\cite{vaswani2017attention, mildenhall2020nerf} $\delta := \theta (\mathbf{x}_i - \mathbf{y}_j)$ that consists of two linear layers with a single ReLU~\cite{nair2010rectified}.
Then, the summation result of $\delta(\mathbf{x}_i - \mathbf{y}_j)$ and $\varphi(\mathbf{g}_j) - \psi(\mathbf{f}_i)$ will be further processed by a MLP $\gamma$. Next, a softmax function $\rho$ is used to generate normalized attention scores that are used to calculate a weighted combination of $\alpha( \mathbf{f}_i) + \delta(\mathbf{x}_i)$ to obtain $\mathbf{f}_i'$.

\begin{figure}[!htb]
    \centering
    \includegraphics[width=\linewidth]{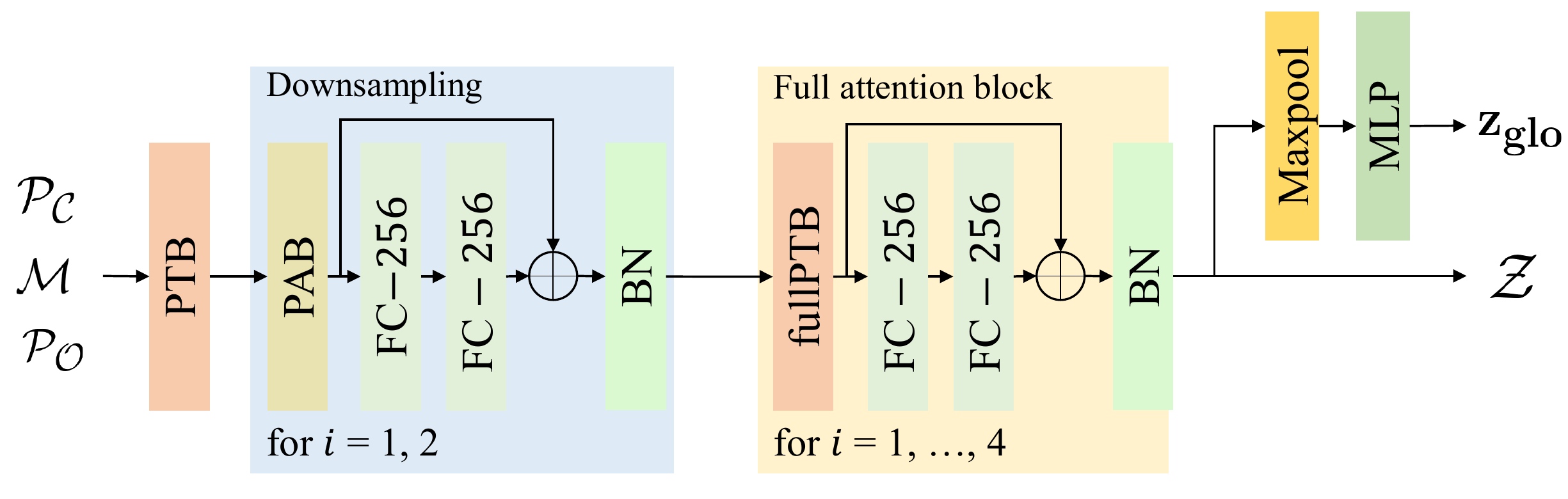}
    \caption{
    \textbf{Point Transformer Encoder}.
    }
    \label{fig:encoder}
\end{figure}

\begin{figure}[!htb]
    \centering
    \includegraphics[width=0.6\linewidth]{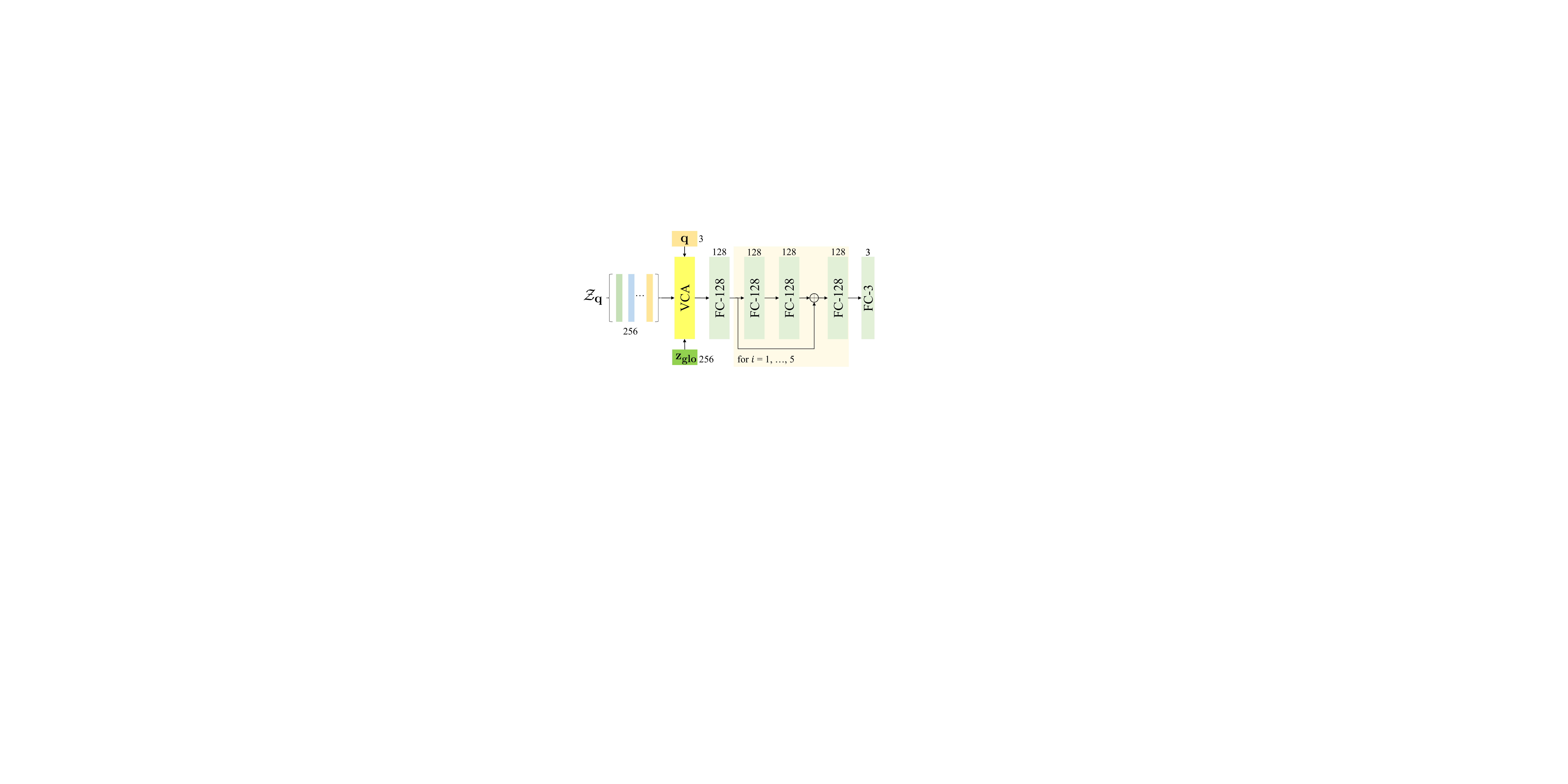}
    \caption{
    \textbf{Attentive Deformation Decoder}.
    }
    \label{fig:decoder}
\end{figure}

\paragraph{Point Transformer Block (PTB):}
As illutrated in \Cref{fig:vca}, we introduce the architecture of point transformer block.
The point transformer block is used to encapsulate the information from $k_{enc}$ = 16 nearest neighborhoods while keeping the position of a point $\mathcal{X}_i$ unchanged.
The input $\mathcal{X}_i$ is fed into a vector attention block (VSA) and through a BatchNorm (BN)~\cite{ioffe2015batch} (including a residual connection from the input $\mathcal{X}_i$).

\paragraph{Point Abstraction Block (PAB):}
The point abstraction block consists of a farthest point sampling module (FPS), a VCA module, a VSA module, followed by a BN layer.
The farthest point sampling (FPS) is used to downsampled $\mathcal{X}$ which is then fed into a VCA followed by a VSA module.
We employ a skip connection from the original $\mathcal{X}$ to the VCA module.
The output of the FPS and the VSA module are fed into a batchnorm layer which computes the output of the point abstraction block.

\paragraph{Point Transformer Encoder}
%
%
As shown in \Cref{fig:encoder}, a PTB is used to obtain an initial feature encoding $\mathcal{Z}_0 = \{ \mathbf{c}_i^0,  \mathbf{z}_i^0\}_{i=1}^{n_0}, n_0=5000$.
Two consecutive point abstraction blocks (PABs) with intermediate set size of $n_1=500$ and $n_2=100$, are used to obtain downsampled feature point clouds $\mathcal{Z}_1 = \{ \mathbf{c}_i^1,  \mathbf{z}_i^1\}_{i=1}^{n_1}$ and $\mathcal{Z}_2 = \{ \mathbf{c}_i^2,  \mathbf{z}_i^2\}_{i=1}^{n_2}$.
To enhance global deformation priors, we stack 4 point transformer block with full self-attention whose $k_{\enc}$ is set to 100 to exchange the global information in the whole set of $\mathcal{Z}_2$.
By doing so, we can obtain a sparse set of local deformation descriptors $\mathcal{Z} = \{ \mathbf{c}_i,  \mathbf{z}_i\}_{i=1}^{100}$ that are anchored in $\{ \mathbf{c}_i \}$.
Finally, a global max-pooling operation followed by two linear layers is used to obtain the global latent vector $\mathbf{z}_{\glo}$.

\paragraph{Attentive Deformation Decoder}
The detailed architecture of attentive deformation decoder is shown in~\Cref{fig:decoder}.
It fuses near-by local latent codes $\mathcal{Z}_{\mathbf{q}}$ of $\mathbf{q}$ under the guidance of a global latent code $\mathbf{z}_\mathbf{glo}$ into $\mathbf{z}$, and feeds $\mathbf{z}$ into
an MLP consisting of five stacked Res-FC blocks to estimate the displacement vector of $\mathbf{q}$.

\section{Evaluation Metrics}
\label{SecEva}


For defining the evaluation metrics, we assume two meshes $\mathcal{T} = \{ \mathcal{V}, \mathcal{F} \}$ and $\mathcal{T}' = \{ \mathcal{V}', \mathcal{F} \}$ being the ground-truth and deformed mesh respectively, sharing the same connectivity.

\paragraph{Vertex $\ell_2$ error:}
The vertex $\ell_2$ distance error is the mean square distance between ground-truth vertices $\mathcal{V} = \{ \mathbf{v}_i \}$ and deformed vertices $\mathcal{V}' = \{ \mathbf{v}_i' \}$:
\[ \ell_2(\mathcal{T}',\mathcal{T}) := \frac{1}{|\mathcal{V}|} \sum_{i=1}^{|\mathcal{V}|} \| \mathbf{v}_i - \mathbf{v}_i'\|_2^2 ,\]
where $|\mathcal{V}|$ denotes the number of mesh vertices.

\paragraph{Chamfer distance:}

To calculate the chamfer distance between $\mathcal{T}'$ and $\mathcal{T}$, we firstly sample two point set $\mathcal{P}_{\mathcal{T}'}$ and $\mathcal{P}_{\mathcal{T}}$ from $\mathcal{T}'$ and $\mathcal{T}$ individually. Then, 
the Chamfer distance of two point sets is defined as:
\[ \CD(\mathcal{T}',\mathcal{T}) := \CD(\mathcal{P}_{\mathcal{T}'}, \mathcal{P}_{\mathcal{T}} ) = \sum_{x\in\mathcal{P}_{\mathcal{T}'}}\min_{y\in\mathcal{P}_{\mathcal{T}}} \|x-y\|_2^2 + \sum_{y\in\mathcal{P}_{\mathcal{T}}}\min_{x\in\mathcal{P}_{\mathcal{T}'}} \|x-y\|_2^2 .\]

\paragraph{Face Normal Consistency}

The face normal consistency describes the mean cosine similarity score of the triangle normals of two meshes.  Let $\mathcal{N}$ and $\mathcal{N}'$ denote the set of face normals of $\mathcal{T}$ and $\mathcal{T}'$ respectively.  We define Face Normal Consistency as:
\[ \FNC(\mathcal{T}',\mathcal{T}) := \frac{1}{|\mathcal{N}|}\sum_{i=1}^{|\mathcal{N}|} |\mathbf{n}' \cdot \mathbf{n} | ,\]
where $|\mathcal{N}|=|\mathcal{F}|$ denotes the number of triangle faces and $\cdot$ denotes the dot product of two vectors.

\section{Notation}
\label{SecNotation}
\rev{
    We will explain our notation in more detail after having briefly defined it in Section 3. By $\mathcal{S}$, $\mathcal{C}$, $\mathcal{T}$, $\mathcal{T}'$ we denote meshes of the considered shapes. $\mathcal{S}=\{\mathcal{V},\mathcal{F}\}$ is the source mesh and $\mathcal{V}$ is the set of vertices of  $\mathcal{S}$ while $\mathcal{F}$ is the set of faces of $\mathcal{S}$. $\mathcal{S}$ is deformed in a 2-step approach. By $\mathcal{C}$ we denote the canonical shape and $\mathcal{T}$ is the target shape.
    We select a sparse set of handles $\mathcal{H} = \{\mathbf{h}_i\}_{i=1}^\ell$ of the original shape. The handles can be dragged to new target locations $\mathcal{O} = \{\mathbf{o}_i\}_{i=1}^\ell$ which define the target mesh $\mathcal{T}$. The continuous deformation field learnt in our work is denoted by $\mathbf{D}$. We apply $\mathbf{D}$ to deform the vertices of $\mathcal{S}$ to obtain the deformed mesh $\mathcal{T}'=\{\mathcal{V}+\mathbf{D}(\mathcal{V}),\mathcal{F}\}$ where $\mathcal{V}+\mathbf{D}(\mathcal{V})$ are the vertices of the deformed mesh. We denote the backward deformation field by $\mathbf{D}_b$ and the forward deformation field by $\mathbf{D}_f$. It holds $\mathbf{D}_f(\mathbf{D}_b(\cdot))$.
    Since our method performs operations in the point cloud domain, we sample point clouds from the surface meshes. $\mathcal{P_S} = \{\mathbf{p}_i\}_{i=0}^n$ is a surface point cloud of canonical mesh $\mathcal{S}$ with size $n=5000$. We define the binary user handle mask as $\mathcal{M} = \{ b_i \;|\; b_i = 1 \text{ if } \mathbf{p}_i \text { is a handle or } b_i = 0 \text{ else, } i=1,\ldots,n\}$. The point cloud $\mathcal{P_S}$ is passed through the backward transformation network $\Omega_b$ and mapped into the canonical pose $\mathcal{P_C'}$, i.e. $\mathcal{P_C'} = \mathcal{P_S} + \mathbf{D}_b(\mathcal{P_S})$. Then the point cloud $\mathcal{P_C'}$ is passed through the forward transformation network $\Omega_f$ and mapped into the target pose $\mathcal{P_T'}$, i.e. $\mathcal{P_T'} = \mathcal{P_C'} + \mathbf{D}_f(\mathcal{P_C'})$.
    Further, consult~\Cref{tab:notation} for the definition of all items.
}
\begin{table*}[h]
	\begin{center}
		\begin{tabular}{cc}
			\toprule
			Notations   & Meaning \\
			 
			\midrule
           $\mathcal{S}, \mathcal{C}, \mathcal{T}, \mathcal{T'}$ & Source mesh, canonical mesh, target mesh, deformed mesh  \\
           $\mathcal{V, F}$  & Vertices, faces of source mesh $\mathcal{S}$ \\
           $\mathcal{H}, \mathbf{h}_i$ & Set of handles, $i$-th handle location\\
           $\mathcal{O}, \mathbf{o}_i$ & Set of target locations of handles, $i$-th target location \\
           $\mathcal{M}, \textbf{b}_i$ & Binary user handle mask, $i$-th element of $\mathcal{M}$ \\
           $\mathcal{P_S}, \mathcal{P_C}, \mathcal{P_T}$ & Surface point clouds of size $n$ sampled from the surface of $\mathcal{S}, \mathcal{C}, \mathcal{T}$ \\
           $\mathcal{P_O}$ & Target handle point locations \\
           $\mathcal{Q_S}, \mathcal{Q_C}, \mathcal{Q_T}$ & Non-surface point clouds of size $m$ sampled from the 3D space of $\mathcal{S}, \mathcal{C}, \mathcal{T}$ \\
           $\mathbf{q}_i$ & $i$-th non-surface querying point \\
           $n$ & Size of surface point clouds $\mathcal{P_S}, \mathcal{P_C}, \mathcal{P_T}$ \\
           $m$ & Size of non-surface point clouds $\mathcal{Q_S}, \mathcal{Q_C}, \mathcal{Q_T}$ \\
           $\mathbf{p}_i$ & $i$-th point from $\mathcal{P_S}$ \\
           $\mathcal{P_C'}, \mathcal{P_T'}$ & Mapping of $\mathcal{P_S}$ in canonical pose, target pose \\
           $\mathbf{D}_b, \mathbf{D}_f$ & Backward deformation field, forward deformation field \\
            $\mathbf{D}$ & Deformation field between two arbitrary poses, i.e. $\mathbf{D}_f(\mathbf{D}_b(\cdot))$\\
           $\Omega_b, \Omega_f$ & Backward transformation network, forward transformation network \\
           $\mathcal{X, Y}$  & Query sequence, key-value sequence \\
           $\mathbf{x}_i, \mathbf{f}_i, \mathbf{f}_i'$ & Coordinate of $i$-th query point, corresponding feature vector, aggregated feature \\
           $\mathbf{y}_j, \mathbf{g}_j$ & Coordinate of $j$-th key-value point, corresponding feature vector \\
           $\VCA$ & Vector cross attention \\
           $\varphi, \psi, \alpha$ & Fully-connected layers \\
           $\gamma$ & Attention weight normalization function, e.g. \textit{softmax} function  \\
           $\delta$ & Positional embedding module \\
           $\VSA$ & Vector self-attention operator \\
           $\PTB, \PAB$ & Point transformer block, point abstraction block \\
           $\BN$ & BatchNorm Layer \\
           $\mathcal{Z}$ & Set of local deformation descriptors \\
           $\mathbf{q}, \mathbf{z_q}$ & A point in $\mathcal{C}$, corresponding feature vector \\
           $\mathbf{c}_i, \mathbf{z}_i$ & Coordinates and feature vector of $i$-th deformation descriptor \\
           $\mathbf{z}_{\glo}$ & Global latent vector \\
           $L_b, L_f, L_{\text{total}}$ & Backward loss function, forward loss function, end-to-end loss function \\
        \bottomrule
        \end{tabular}
        \caption{\rev{Notations in order of appearance in the main paper.}}
        \label{tab:notation}
        \end{center}
\end{table*}

\section{Data}
\label{SecData}

To train and evaluate our method, we use the DeformingThing4D~\cite{li20214dcomplete} dataset, which is available under a non-commercial academic license.
It does not contain personally identifiable information or offensive contents.
We have obtained the consent to use the dataset.
\rev{
\paragraph{Train/test split}
The DeformingThing4D consists of a large number of quadruped animal animations with various motions, 
such as “bear3EP Jump”, “bear9AK Jump”, or “bear3EP Lie” where "bear3EP" and "bear9AK" are identity names, and "Jump" and "Lie" are motion names.
Similar to the D-FAUST~\cite{bogo2017dynamic} used in OFlow~\cite{niemeyer2019occupancy}, the train/test split is based on these identity and motion names of deforming sequences.
We firstly divide the animations of the dataset into two parts, seen identities and unseen identities. 
For the animations of seen identities, we further divide it into seen motions of seen identities (used as training set), and unseen motions of seen identities (used as the test set of S1).
The animations of unseen identities are used as the test set of S2.
%
Finally, the train, test S1, and test S2 datasets individually contains 1296, 143, and 55 deforming sequences.
\paragraph{Data preparation}
In Section 3.3 of the main text, we mentioned that  our method utilizes a set of triplets including source $\mathcal{S}$, canonical $\mathcal{C}$, and target mesh $\mathcal{T}$ with dense correspondence for training. The point clouds $\mathcal{P}_\mathcal{S}, \mathcal{P}_\mathcal{C}, \mathcal{P}_\mathcal{T}$ of size $n$ with one-to-one correspondence are sampled from the surfaces of $\mathcal{S}$, $\mathcal{C}$, $\mathcal{T}$.  And the non-surface point sets $\mathcal{Q}_\mathcal{S}, \mathcal{Q}_\mathcal{C}, \mathcal{Q}_\mathcal{T}$ of size $m$  are sampled from their 3D space. Here, we provide the details of data preparation. 
Firstly, we sample $N_p$ surface points $\{ \mathbf{x}_i \}_{i=1}^{i=N_p}$ from the canonical
mesh $\mathcal{C}$; we also store the corresponding barycentric weights of sample points. Then, each point is randomly permuted by a small displacement vector $\delta_{\mathbf{n}_i} = \mathbf{x}_i + \beta * \mathbf{n}_i$ along the normal direction $\mathbf{n}_i$ of the corresponding triangle. The displacement distance $\beta$ is from a Gaussian distribution $N(0, \sigma^2)$. Next, for source $\mathcal{S}$ and target  $\mathcal{T}$ meshes, we use the same barycentric weights to obtain $\mathcal{P}_\mathcal{S}, \mathcal{P}_\mathcal{T}$ with correspondences, and use the same displacements $\delta_{\mathbf{n}}$ to obtain $\mathcal{Q}_\mathcal{S}, \mathcal{Q}_\mathcal{T}$ with correspondences.
Concretely, we pre-compute $N_p=20{,}000$ points from each canonical surface mesh, and get the non-surface points with 50\% of surface points permuted by $\sigma=0.02$, with 50\% of surface points permuted by $\sigma=0.1$. During training, we down-sample $n=5000$ points of $\mathcal{P}_\mathcal{S}, \mathcal{P}_\mathcal{C}, \mathcal{P}_\mathcal{T}$,
and down-sample $m=5000$ of $\mathcal{Q}_\mathcal{S}, \mathcal{Q}_\mathcal{C}, \mathcal{Q}_\mathcal{T}$. To maintain one-to-one correspondence, we use the same sampling indices for $\mathcal{S}$, $\mathcal{C}$, $\mathcal{T}$.
}

\section{Additional Results}
\label{SecAdd}

%
\paragraph{Effects of point cloud sampling density}
To study the effect of sampling density of input point cloud, we individually train our model by using point clouds of size 2500, 5000, 7500 as input.
Quantitative results are shown in~\Cref{tab:point_density}.
We can observe that the results of different evaluation metrics only show a slightly small variance.
To balance accuracy and computational cost, we use 5000 points in our final model.
\begin{table*}[h]
	\renewcommand\arraystretch{1.2}
	\begin{center}
		\begin{tabular}{*{7}{c}}
			\toprule
			\multirow{2}*{\#sampling points} & \multicolumn{3}{c}{New motions (S1)}   & \multicolumn{3}{c}{Unseen identities (S2)} \\
			\cmidrule(lr){2-4} \cmidrule(lr){5-7}
			 & $\ell_2$ $\downarrow$ & CD $\downarrow$ & FNC $\uparrow$   & $\ell_2$ $\downarrow$ & CD $\downarrow$ & FNC $\uparrow$  \\
			\midrule
			\midrule
            Ours-2500
                   & 0.789 & 1.008 & 96.27
                   & 0.905 & 1.285 & 96.57 \\

            Ours-5000 
                   & 0.752 & 0.948  & \textbf{96.59} 
                   & 0.795 & \textbf{1.241}  & \textbf{96.68}  \\

            Ours-7500
                   & \textbf{0.732} & \textbf{0.944} & 96.39
                   & \textbf{0.789} & 1.251 & 96.66  \\

        \bottomrule
        \end{tabular}
        \caption{Quantitative results of different input point cloud density on the S1 and S2 test sets of DeformingThing4D~\cite{li20214dcomplete} dataset. 
        }
        \label{tab:point_density}
        \end{center}
\end{table*}

%
%
\paragraph{Robustness to noisy source mesh}
\rev{
To analyze the robustness of noise effects,
we individually train our model by adding gaussian noise permutations to the source meshes.
The standard deviation of gaussian noise is set to 0, 0.0025 or 0.005.
The comparison in~\Cref{tab:noise_level} shows that with the noise becoming larger, the performance of our method experiences only slight variation; however, this demonstrates the robustness of our method to noisy source meshes.
}

\begin{table*}[h]
	\renewcommand\arraystretch{1.2}
	\begin{center}
		\begin{tabular}{*{7}{c}}
			\toprule
			\multirow{2}*{\#standard deviation} & \multicolumn{3}{c}{New motions (S1)}   & \multicolumn{3}{c}{Unseen identities (S2)} \\
			\cmidrule(lr){2-4} \cmidrule(lr){5-7}
			 & $\ell_2$ $\downarrow$ & CD $\downarrow$ & FNC $\uparrow$   & $\ell_2$ $\downarrow$ & CD $\downarrow$ & FNC $\uparrow$  \\
			\midrule
			\midrule
            Ours-0
                    & \textbf{0.752} & \textbf{0.948}  & \textbf{96.59} 
                    & \textbf{0.795} & \textbf{1.241}  & \textbf{96.68}  \\

            Ours-0.0025
                   & 0.774 & 0.973  & 95.90
                   & 0.808 & 1.278  & 96.65  \\

            Ours-0.0050
                   & 0.851 & 1.017 & 96.50
                   & 0.911 & 1.392 & 96.16  \\

        \bottomrule
        \end{tabular}
        \caption{Quantitative results of source meshes with different noise intensities on the S1 and S2 test sets of DeformingThing4D~\cite{li20214dcomplete} dataset. 
        }
        \label{tab:noise_level}
        \end{center}
\end{table*}

\paragraph{Robustness to partial source mesh}
\rev{
To investigate the robustness to incomplete source meshes, we randomly sample 5 seeds from the source mesh surface, and then remove the corresponding $k_r$ nearest vertices and corresponding faces. The $k_r$ is calculated by $k_r=p_r*|\mathcal{V}|$, where $p_r$ is the incompleteness ratio and $ |\mathcal{V}|$ is the number of source mesh vertices. Again, our model is directly evaluated under two different settings of $p_r=0.05$ and $p_r =0.1$. The quantitative results are provided in ~\Cref{tab:partial_input}.
As seen, there are not significant numerical variations between different  incompleteness ratios.
This clearly demonstrates the robustness of our approach to incomplete source meshes.
}

\begin{table*}[h]
	\renewcommand\arraystretch{1.2}
	\begin{center}
		\begin{tabular}{*{7}{c}}
			\toprule
			\multirow{2}*{\#incompleteness ratio} & \multicolumn{3}{c}{New motions (S1)}   & \multicolumn{3}{c}{Unseen identities (S2)} \\
			\cmidrule(lr){2-4} \cmidrule(lr){5-7}
			 & $\ell_2$ $\downarrow$ & CD $\downarrow$ & FNC $\uparrow$   & $\ell_2$ $\downarrow$ & CD $\downarrow$ & FNC $\uparrow$  \\
			\midrule
			\midrule
            Ours-0.0
                   & \textbf{0.752} & \textbf{0.948}  & \textbf{96.59} 
                   & \textbf{0.795} & \textbf{1.241}  & \textbf{96.68}  \\

            Ours-0.05
                  & 0.770  & 0.957 & 95.80  
                  & 0.804  & 1.244 & 96.66 \\
            
            Ours-0.10
                   & 0.823 & 1.002  & 96.44 
                   & 0.858 & 1.261  & 96.55  \\
                  %

        \bottomrule
        \end{tabular}
        \caption{Quantitative results of source meshes with different incomplete ratios on the S1 and S2 test sets of DeformingThing4D~\cite{li20214dcomplete} dataset. 
        \rev{Note that our model is directly evaluated on partial meshes without fine-tuning.}
        }
        \label{tab:partial_input}
        \end{center}
\end{table*}

\begin{figure}[h]
        \centering
        \includegraphics[width=\linewidth]{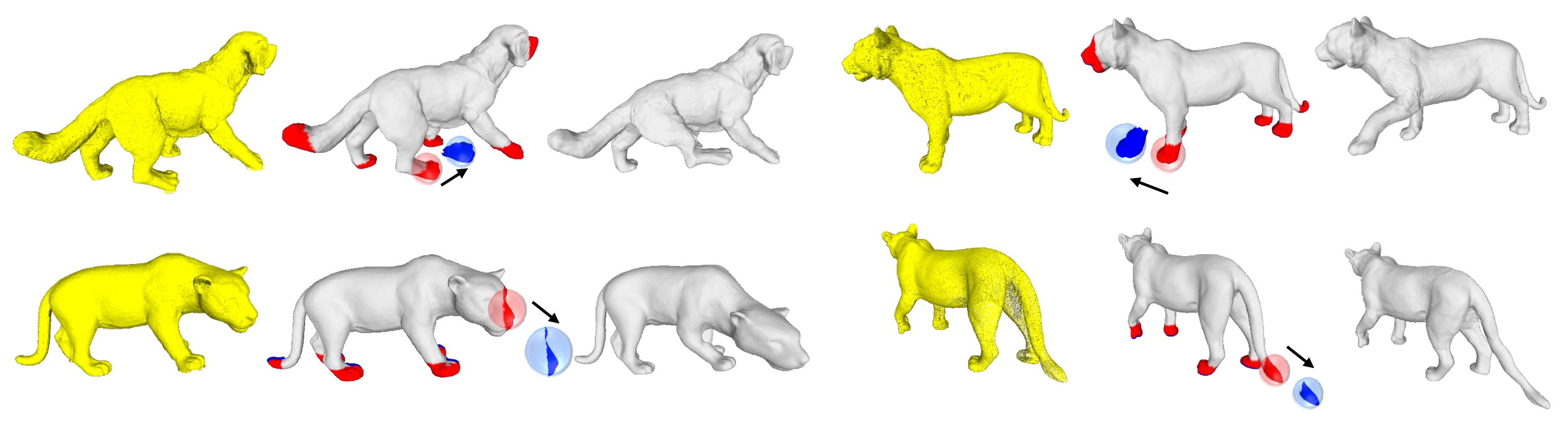}
        \begin{tabular}{@{\hspace*{3.5em}}p{60pt}p{55pt}p{55pt}p{55pt}p{45pt}p{40pt}}
                  (a) & 
                  (b) &
                  (c) &
                  (a) &
                  (b) &
                  (c) 
        \end{tabular}
        \caption{
        \rev{
        Evaluation on real animal scans. (a) Real animal scans  (b) Source meshes obtained via the Screened PSR~\cite{kazhdan2013screened} and handles. (c) Ours.}
        }
        \label{fig:real_scan}
    \end{figure}

\begin{figure}[h]
        \centering
        \includegraphics[width=\linewidth]{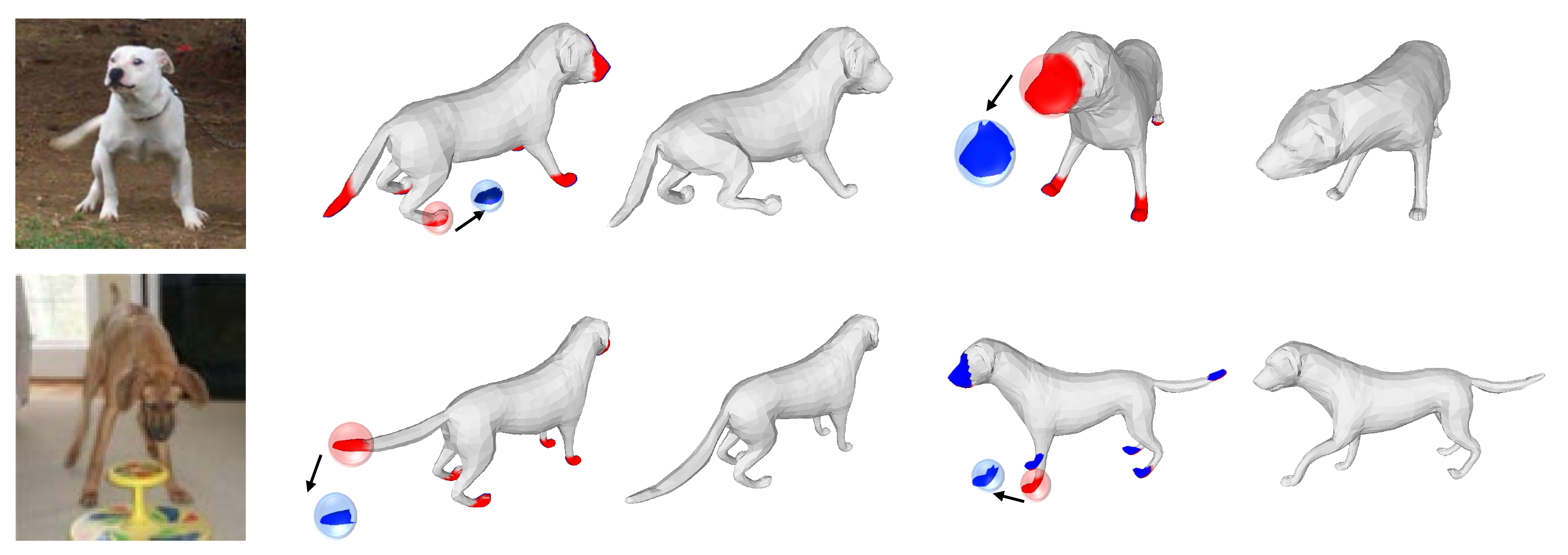}
        \begin{tabular}{@{\hspace*{3.5em}}p{75pt}p{65pt}p{60pt}p{50pt}p{50pt}}
                  (a) & 
                  (b) &
                  (c) &
                  (d) &
                  (e)
        \end{tabular}
        \caption{
        \rev{
        Evaluation on reconstructed animals from real RGB images using the method of BARC~\cite{ruegg2022barc} (a) Real images. (b) Reconstructed source meshes and handles. (c) Ours. (d) Reconstructed source meshes and handles. (e) Ours.}
        }
        \label{fig:barc_dogrec}
    \end{figure}

\begin{figure}[h]
        \centering
        \includegraphics[width=\linewidth]{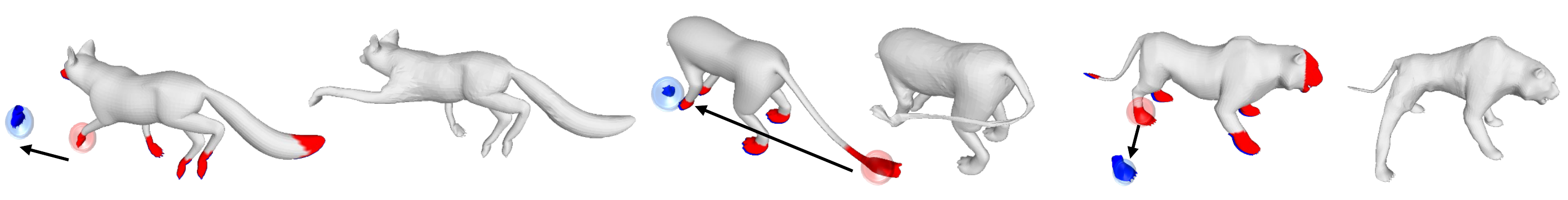}
        \begin{tabular}{@{\hspace*{3.5em}}p{60pt}p{60pt}p{45pt}p{45pt}p{50pt}p{55pt}}
                  (a) & 
                  (b) &
                  (a) &
                  (b) &
                  (a) &
                  (b) 
        \end{tabular}
        \caption{
        \rev{
        Evaluation on non-realistic user-specified handles. (a) Source meshes and handles. (b) Ours.}
        }
        \label{fig:nonrealistic}
    \end{figure}

\begin{figure}[!hbt]
        \centering
        \includegraphics[width=\linewidth]{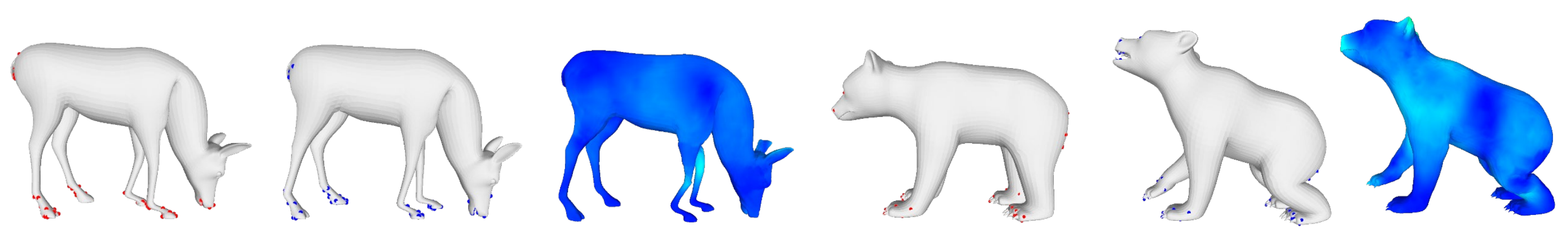}
        \begin{tabular}{@{\hspace*{3.5em}}p{50pt}p{50pt}p{60pt}p{50pt}p{55pt}p{55pt}}
                  (a) & 
                  (b) &
                  (c) &
                  (a) & 
                  (b) &
                  (c) 
        \end{tabular}
        \caption{
            \rev{
            The evaluation results of our modified framework that uses sparse handles as input and does not require dense correspondences as supervision. (a) Source meshes and handles. (b) Target meshes and handles. (c) Our results with vertex error map.
            }
        }
        \label{fig:sparsehandles_chamfer}
    \end{figure}

\paragraph{Evaluations on real animals scans.}
\rev{
We evaluate our pre-trained model on the real animal scans captured by ourselves. As show in \Cref{fig:real_scan}, our method can still learn realistic shape deformations, which demonstrates the generalization ability of our approach to real captured models.}
\paragraph{Evaluations on reconstructed animals from real images.}
    In addtion, we evaluate our pre-trained model on the reconstructed animals from real RGB images using the BARC~\cite{ruegg2022barc} method. As shown in~\Cref{fig:barc_dogrec}, our method estimates realistic deformations for reconstructed animals from natural images. This also demonstrates the generalization ability of our method.
\paragraph{Evaluations on non-realistic user-specified handles.}
\rev{ While our goal of data-driven deformation priors is to obtain deformations that are as realistic as possible, we also evaluate our method on non-realistic or non-physical-aware handles. As shown in~\Cref{fig:nonrealistic}, our method will try to find the closest deformation of animals that can best explain the provided handle displacements.
However, our method could be easily trained on non-realistic or non-physical-aware samples and learn the respective deformation behavior.
}
\paragraph{Without dense correspondence}
While our current method uses an existing dataset where dense correspondences between temporal mesh frames are available, our framework can also be trained on datasets without dense correspondences through some adjustments on inputs and loss functions. Concretely, we change our method to receive sparse handle correspondences as inputs, and utilize Chamfer distance as the loss function that does not require ground-truth meshes with dense correspondences as supervision. In ~\Cref{fig:sparsehandles_chamfer}, we visualize several test results of such a modified framework. As seen, without dense correspondences for training, our method can still obtain accurate deformations.
\paragraph{Video animations}
To visualize the deformation behaviours of the different approaches, we use a sequence of handle movements as inputs, and run our model frame by frame to obtain a deformation motion sequence.
We refer to the supplemental video for an animated sequence.
\section{Limitations}
\label{SecLim}
    \begin{figure}[!hbt]
        \centering
        \includegraphics[width=\linewidth]{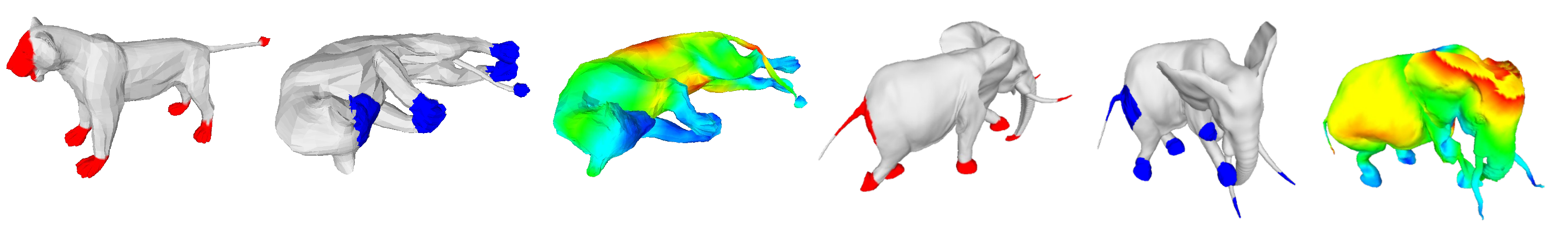}
        \begin{tabular}{@{\hspace*{3.5em}}p{50pt}p{50pt}p{60pt}p{50pt}p{55pt}p{55pt}}
                  (a) & 
                  (b) &
                  (c) &
                  (a) & 
                  (b) &
                  (c) 
        \end{tabular}
        \caption{The failure cases. (a) Source meshes and handles. (b) Target meshes and handles.
                  (c) Our results with vertex error map.}
        \label{fig:fail}
    \end{figure}
While compelling results have been demonstrated for shape manipulation, a few limitations still exist in our approach that can be addressed in future work.
Two representative failure cases are depicted in~\Cref{fig:fail}.
We can see that our method cannot well address extreme shape deformations (e.g. left of~\Cref{fig:fail}) or manipulate unseen identities that are far from the training data distribution (e.g. the elephant in the right of~\Cref{fig:fail}).
We believe this issue can be alleviated by a larger training dataset, a richer data augmentation strategy, and/or few shot generalization techniques in the future.

\end{document}